\documentclass[11pt]{article}

\usepackage[final]{acl}

\usepackage{times}
\usepackage{latexsym}

\usepackage[T1]{fontenc}

\usepackage[utf8]{inputenc}

\usepackage{microtype}

\usepackage{inconsolata}

\usepackage{booktabs}
\usepackage{multirow}
\usepackage{graphicx}
\usepackage[table]{xcolor}
\usepackage{amssymb}
\usepackage{amsmath}

\usepackage{float}
\usepackage{pgfplots}
\pgfplotsset{compat=1.18}

\usepackage{makecell} 

\providecommand{\tabcheck}{\textcolor{teal!70!black}{$\checkmark$}}
\providecommand{\tabbest}[1]{\textbf{#1}}
\providecommand{\tabsecond}[1]{\underline{#1}}
\newcommand{\dpos}[1]{\textcolor{green!45!black}{\scriptsize(+#1)}}
\newcommand{\dneg}[1]{\textcolor{red!70!black}{\scriptsize($-$#1)}}
\newcommand{\vc}[2]{#1\,{\scalebox{0.85}{#2}}}

\usepackage[most]{tcolorbox}
\usepackage{listings}

\definecolor{promptgreen}{HTML}{A8C9AD}
\definecolor{promptbackground}{HTML}{F7FAF7}
\definecolor{carepink}{HTML}{EB7371}

\lstdefinestyle{promptstyle}{
    basicstyle=\ttfamily\scriptsize,
    breaklines=true,
    columns=fullflexible,
    keepspaces=true,
    showstringspaces=false
}

\newtcblisting{promptbox}[1]{
    enhanced,
    breakable,
    listing only,
    title={#1},
    colback=promptbackground,
    colframe=promptgreen,
    colbacktitle=promptgreen,
    coltitle=white,
    fonttitle=\bfseries,
    boxrule=0.8pt,
    arc=2mm,
    left=1.5mm,
    right=1.5mm,
    top=1.5mm,
    bottom=1.5mm,
    before skip=6pt,
    after skip=8pt,
    listing options={style=promptstyle}
}

\title{CARE-3D: Reinforcement Learning for Evidence Grounded Agentic Reasoning in Situated 3D Scenes}

\author{Xiaoqian Hu, Ruihong Qiu, Yanran Tang, Sen Wang \\
  School of Electrical Engineering and Computer Science \\
  The University of Queensland \\
  Brisbane, Queensland, Australia \\
  \texttt{\{xiaoqian.hu, r.qiu, yanran.tang, sen.wang\}@uq.edu.au}}

\begin{document}
\maketitle

\begin{abstract}
Situated 3D reasoning requires embodied agents to interpret objects, scene geometry, and viewpoint dependent spatial relations. Existing methods learn grounded interactions from supervised traces, but largely imitate prescribed evidence acquisition patterns rather than optimizing the support that the returned evidence provides for the correct answer. Moreover, outcome rewards provide limited credit for intermediate tool choices and may leave infrequent tools underexplored. An effective agent should instead adaptively decide what to ground, which tool to invoke, and how to assess the utility of returned evidence. To this end, this paper presents CARE-3D, a reinforcement post-training framework for adaptive grounded reasoning. Supervised fine-tuning (SFT) on grounding instructions first establishes valid interactions, after which Group Relative Policy Optimization (GRPO) optimizes the agent while keeping the 3D grounding module frozen. CARE-3D combines Counterfactual Attribution of Returned Evidence for evidence sensitive credit assignment, and Group Coverage Sampling to preserve tool coverage. Experiments on MSQA and Beacon3D show consistent gains over SFT and GRPO using only answer and format rewards, especially in spatial reasoning and grounding answer coherence. Our code is available at \href{https://github.com/xiaoqian-UQ/CARE-3D}{CARE-3D}.

\end{abstract}

\section{Introduction}

Understanding and reasoning about 3D environments is a fundamental capability for intelligent agents operating in the physical world. Effective 3D reasoning requires agents to recognize objects, interpret scene geometry, and connect language with their surroundings~\cite{3dllm,LEO}. Situated 3D reasoning further requires these relations to be understood from a specific observer position and orientation, making spatial meaning dependent on where the agent is located and facing~\cite{sqa3d,msqa}. This viewpoint aware capability supports embodied question answering, navigation, vision-language navigation, and grounded instruction following. It therefore represents an important step toward reliable and general purpose embodied intelligence.

\begin{figure}[!t]
    \centering
    \includegraphics[width=\linewidth]{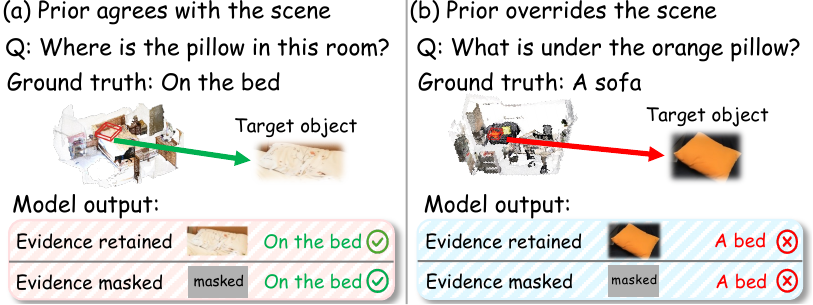}
    \caption{Evidence-invariant predictions of the SFT policy. Predictions remain unchanged when the returned crop is replaced with a fixed placeholder. The pillow-bed association matches the scene in (a) but conflicts with it in (b), illustrating why answer correctness alone does not establish evidence use.}
    \label{fig:motivation}
\end{figure}

Existing 3D vision-language research has progressed from global scene representation toward explicit and situated grounding. Here, \emph{grounding} refers to identifying the 3D scene objects referred by a question or grounding query. Scene level models encode point clouds, object centric features, or multi-view observations to support semantic and spatial reasoning~\cite{3dllm, pointllm, LEO}. Subsequent methods introduce explicit object references and 3D positional information to connect language reasoning with scene entities~\cite{grounded3dllmreferenttokens, ChatScene, llava3d}. Situated approaches further incorporate the observer's position and orientation~\cite{Yuan_2025_CVPR}, while SceneCOT uses supervised grounding traces to acquire task specific scene evidence before answering~\cite{scenecot}. These advances establish increasingly explicit and viewpoint aware interactions between language models and 3D environments.

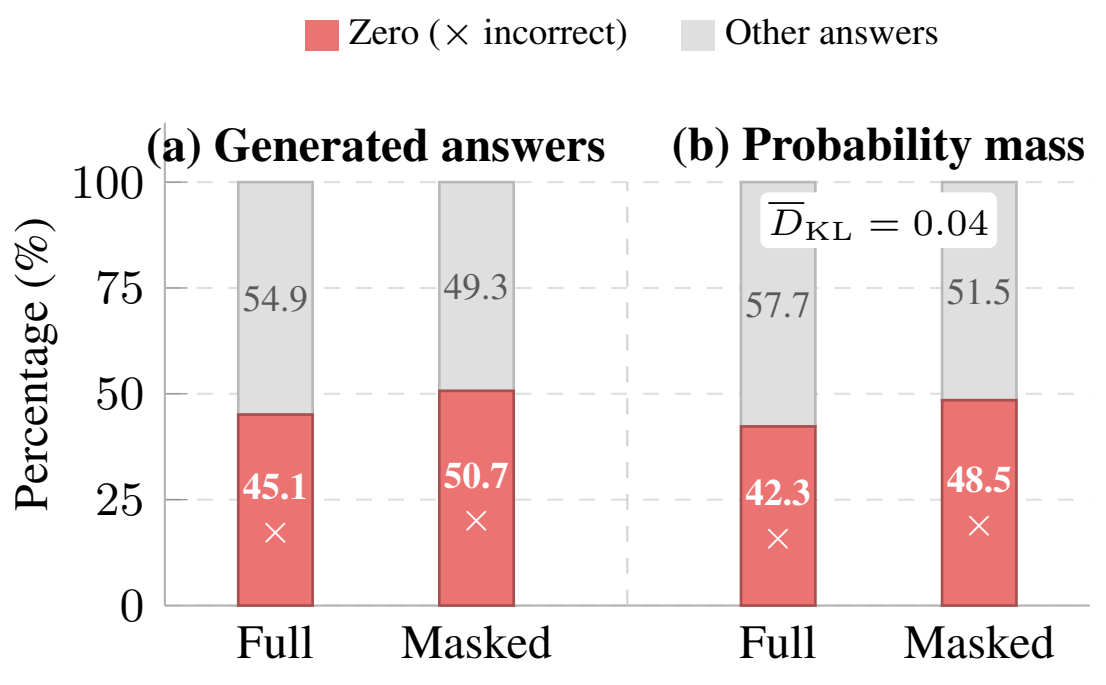
\begin{figure}[!t]
\centering
\begin{tikzpicture}
\begin{axis}[
    set layers,
    ybar stacked,
    width=\columnwidth,
    height=0.62\columnwidth,
    xmin=-0.55,
    xmax=4.05,
    ymin=0,
    ymax=114,
    bar width=14pt,
    ylabel={Percentage (\%)},
    ylabel style={font=\footnotesize, yshift=-5pt},
    ytick={0,25,50,75,100},
    xtick={0,1,2.5,3.5},
    xticklabels={Full,Masked,Full,Masked},
    x tick label style={font=\footnotesize},
    tick label style={font=\footnotesize},
    axis x line*=bottom,
    axis y line*=left,
    axis line style={gray!60},
    ymajorgrids,
    grid style={dashed,gray!25},
    clip=false,
    legend style={
        at={(0.5,1.10)},
        anchor=south,
        legend columns=2,
        draw=none,
        fill=none,
        font=\scriptsize,
        /tikz/every even column/.append style={column sep=8pt}
    },
    legend image code/.code={
        \draw[#1,draw=none]
        (0cm,-0.1cm) rectangle (0.22cm,0.1cm);
    },
]

\addplot[
    fill=carepink,
    draw=carepink!70!black,
    line width=0.5pt
] coordinates {
    (0,45.1)
    (1,50.7)
    (2.5,42.3)
    (3.5,48.5)
};
\addlegendentry{Zero ($\times$ incorrect)}

\addplot[
    fill=gray!25,
    draw=gray!55,
    line width=0.5pt
] coordinates {
    (0,54.9)
    (1,49.3)
    (2.5,57.7)
    (3.5,51.5)
};
\addlegendentry{Other answers}

\node[
    on layer=axis foreground,
    font=\footnotesize\bfseries
] at (axis cs:0.5,108)
    {(a) Generated answers};

\node[
    on layer=axis foreground,
    font=\footnotesize\bfseries
] at (axis cs:3.0,108)
    {(b) Probability mass};

\pgfplotsextra{
  \begin{pgfonlayer}{axis foreground}

    \node[
        font=\scriptsize,
        fill=white,
        rounded corners=1.5pt,
        inner sep=1.5pt
    ] at (axis cs:3.0,91)
        {$\overline{D}_{\mathrm{KL}}=0.04$};

    \node[font=\scriptsize\bfseries,white,align=center]
        at (axis cs:0,22.55) {45.1\\$\times$};
    \node[font=\scriptsize\bfseries,white,align=center]
        at (axis cs:1,25.35) {50.7\\$\times$};
    \node[font=\scriptsize\bfseries,white,align=center]
        at (axis cs:2.5,21.15) {42.3\\$\times$};
    \node[font=\scriptsize\bfseries,white,align=center]
        at (axis cs:3.5,24.25) {48.5\\$\times$};

    \node[font=\scriptsize,text=gray!70!black]
        at (axis cs:0,72.55) {54.9};
    \node[font=\scriptsize,text=gray!70!black]
        at (axis cs:1,75.35) {49.3};
    \node[font=\scriptsize,text=gray!70!black]
        at (axis cs:2.5,71.15) {57.7};
    \node[font=\scriptsize,text=gray!70!black]
        at (axis cs:3.5,74.25) {51.5};

  \end{pgfonlayer}
}

\draw[
    on layer=axis foreground,
    gray!35,
    dashed
] (axis cs:1.75,0) -- (axis cs:1.75,100);

\end{axis}
\end{tikzpicture}

\caption{
Evidence insensitive zero-count bias of the SFT policy on 71 MSQA counting errors with non-zero targets. \emph{Full} retains the returned evidence, whereas \emph{Masked} replaces it with a fixed placeholder. The similar (a) zero-answer frequencies and (b) probability masses, together with the low mean divergence ($\overline{D}_{\mathrm{KL}}=0.04$), indicate limited sensitivity to grounding evidence.
}

\label{fig:language_prior}
\end{figure}

Despite this progress, supervised grounding traces do not guarantee effective evidence use. Models may produce plausible answers despite incorrect grounding, revealing weak grounding answer coherence~\cite{Beacon3D}. Because each trace prescribes only one grounding call, it neither evaluates alternative tools nor verifies that the returned evidence supports the answer. Figure~\ref{fig:motivation} illustrates this issue by comparing two evidence settings: \emph{Full}, which retains the evidence returned by the grounding tool, and \emph{Masked}, which replaces that evidence with a fixed placeholder while keeping the question and grounding request unchanged. The SFT policy produces the same prediction in both settings, succeeding when its pillow-bed prior matches the scene but failing when the pillow is on a sofa.
Answer level rewards therefore cannot reliably distinguish evidence grounded reasoning from predictions driven by textual priors~\cite{IGPO,infopo}. Figure~\ref{fig:language_prior} further shows that, on 71 counting errors with non-zero ground truth answers, the policy predicts \emph{zero} in nearly half the cases and changes little after evidence masking. Meanwhile, on-policy reinforcement learning may reduce sampling diversity, causing low frequency tools to disappear before their utility is evaluated~\cite{dapo,yue2025does}. These limitations motivate evidence sensitive credit assignment and sustained grounding tool coverage.

This paper presents CARE-3D, a GRPO based reinforcement post-training framework that treats grounding as an adaptive information acquisition decision~\cite{DeepSeekMath}. CARE estimates evidence utility from the change in ground truth answer likelihood after masking the returned evidence. GCS adds loss masked rollouts for underrepresented tools, allowing evidence returned by these tools to participate in group relative comparison. Experiments on MSQA~\cite{msqa} and Beacon3D~\cite{Beacon3D} show consistent gains over SFT and answer and format only GRPO.

The main contributions are:
\begin{itemize}
    \item This paper formulates situated 3D grounding as adaptive evidence acquisition and presents CARE-3D, a GRPO based reinforcement post-training framework.

    \item CARE is introduced for evidence sensitive credit assignment and GCS for preserving grounding tool
    coverage during optimization.

    \item Experiments on MSQA and Beacon3D show consistent gains over SFT and GRPO using only answers and format rewards,
    with ablations validating both components.
\end{itemize}

\begin{figure*}[t]
    \centering
    \includegraphics[width=\linewidth]{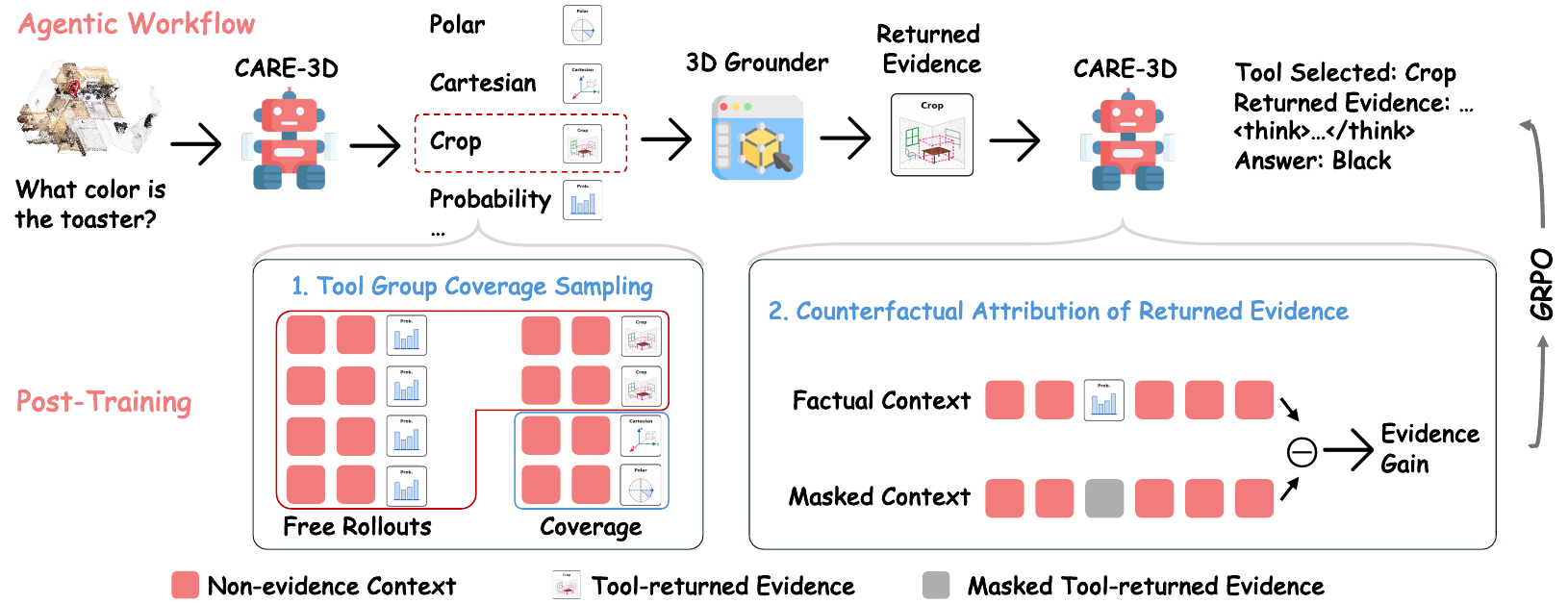}
    \caption{Overview of CARE-3D. At inference, the agent selects a grounding tool and reasons over its returned evidence. During post-training, GCS supplements free rollouts with underrepresented tools, while CARE computes Evidence Gain by contrasting ground truth answer likelihood under factual and evidence masked contexts. GRPO updates the agent using the resulting rewards.}
    \label{fig:framework}
\end{figure*}

\section{Related Work}
\paragraph{3D Vision-Language Reasoning.} 3D vision-language models use different scene representations to connect language reasoning with 3D environments. 3D-LLM aligns LLMs with multi-view 3D features, while PointLLM directly instruction tunes language models over point cloud inputs~\cite{3dllm,pointllm}. LEO represents scenes through object centric tokens to support embodied perception, reasoning, and action~\cite{LEO}. Grounded 3D-LLM introduces referent tokens for linking generated language to scene objects, whereas Chat-Scene and LLaVA-3D incorporate object level scene representations and 3D positional information~\cite{grounded3dllmreferenttokens,ChatScene,llava3d}. MSQA evaluates viewpoint dependent situated reasoning, while Beacon3D examines grounding answer coherence~\cite{msqa,Beacon3D}. SceneCOT further constructs supervised grounding traces for stepwise evidence acquisition~\cite{scenecot}. Unlike these approaches, CARE-3D reinforcement post trains grounding decisions according to the utility of the evidence they return.

\paragraph{Grounded and Tool Augmented 3D Agents.} 
Tool augmented 3D agents invoke external perception and geometry modules to acquire task specific scene evidence. Earlier visual programming methods use LLM generated executable programs to compose off the shelf vision modules for training free visual reasoning~\cite{VISPROG}. Think3D and RieMind coordinate tools such as object detection, depth estimation, and 3D reconstruction for spatial reasoning~\cite{RieMind,think3d}. Skill-3D further learns scene aware skills from multi tool interaction trajectories~\cite{skill3d}. These methods establish effective interfaces between language agents and 3D perception modules, but mainly learn tool use from constructed trajectories or aggregate task outcomes. CARE-3D focuses on directly attributing answer level utility to the evidence returned by each grounding call.

\paragraph{Reinforcement Post-Training and Credit Assignment.}
Reinforcement post-training has extended from language reasoning to multimodal and tool using agents. Visual-RFT applies GRPO with task specific visual rewards, while DeepEyes and VTool-R1 train VLMs to interleave reasoning with visual tool calls~\cite{visual-rtf,deepeyes,VTool-R1}. For intermediate credit assignment, IGPO measures the increase in correct answer probability after each interaction, while InfoPO compares subsequent action distributions under factual and masked feedback~\cite{IGPO, infopo}. TACO assigns credit to individual tool calls using differential answer probes and routes outcome credit to the responsible interaction segments~\cite{taco}. These methods motivate deriving process credit from policy changes without an external reward model. CARE instead attributes the utility of multimodal evidence returned by a 3D grounding call through counterfactual ground truth answer likelihood.

\section{Problem Setup}
\label{sec:problem_setup}
Situated 3D reasoning requires an agent to answer a natural language question $q$ about a 3D scene $\mathcal{S}$ from a specific observer viewpoint $v$, optionally accompanied by a situation description $s$. The viewpoint is essential because spatial expressions such as \emph{left}, \emph{behind}, and \emph{two o'clock} depend on the observer's position and orientation. Each task instance therefore consists of $(\mathcal{S},v,s,q)$, while the agent initially receives the textual input $x=(s,q)$ and accesses the scene and viewpoint through a 3D grounding module. Since the scene may contain many irrelevant objects and relations, the agent must identify question relevant scene entities and acquire appropriate grounded evidence $e$ before answering. Based on this evidence, the agent generates $\hat{y}\sim\pi_{\theta}(\cdot\mid x,a^{\mathrm{grd}},e)$, where $\pi_{\theta}$ denotes the agent policy and $y^\star$ denotes the ground truth answer. The objective is to produce an answer $\hat{y}$ that is semantically consistent with $y^\star$ and supported by both the 3D scene and the observer viewpoint.

\section{Method}
CARE-3D optimizes how a situated 3D reasoning agent acquires and uses scene evidence, as illustrated in Figure~\ref{fig:framework}. The agent is first initialized through supervised grounding instruction tuning following SceneCOT~\cite{scenecot}. The resulting grounding module is then frozen, while GRPO optimizes the agent's grounding requests, tool use, reasoning, and answers. CARE-3D extends standard GRPO with Group Coverage Sampling (GCS) for diversified tool use and Counterfactual Attribution of Returned Evidence (CARE) for evidence sensitive credit assignment.

\subsection{Input Template}
\label{sec:input_template}
Given the textual input \(x=(s,q)\) defined in Section~\ref{sec:problem_setup}, the agent first generates a grounding request with a referential query and a tool identifier. The selected tool returns evidence from the 3D scene. In the second turn, the agent reasons over this evidence and produces its final answer. Box~1 specifies the output format for both turns and refers to the returned evidence as an \emph{observation}.

\begin{promptbox}{Box 1: System Prompt}
You are a precise 3D scene assistant.
You should obtain one observation before answering.
First assistant turn must contain exactly:
<think_grd>evidence query</think_grd> tool from Tool list.
The query line must be a short noun phrase from the current question. Use a concrete object, relation, viewpoint cue, or evidence phrase.
Then output one tool tag from Tool list.
1. <list_obj_prob> Returns ranked object/category evidence with confidence scores.
2. <list_obj_loc_prob> Returns ranked object evidence with scene coordinate 3D centers, sizes, and prob.
3. <list_obj_loc_plr_prob> Returns ranked object evidence with viewer relative angle_deg,
distance, and prob.
4. <highlight_obj> Returns visual evidence for queried object.
Numeric interpretation:
- prob: higher means more likely to match the query.
- <list_obj_loc_prob>: [center_x, center_y, center_z, size_x, size_y, size_z]. Centers are scene coordinate positions. Sizes are extents along each axis.
- <list_obj_loc_plr_prob>: object_name: angle_deg,distance; prob: confidence. The first number is angle_deg, and the second number is distance. angle_deg is relative to viewer forward: 0 front, positive left, negative right, +/-180 behind. clock hour = round(((-angle_deg) mod 360) / 30),  with 0 mapped to 12.
Choose one tool from Tool list according to what returned evidence would make the answer best supported.
After receiving the observation, answer:
<reasoning> brief evidence based rationale </reasoning>
<answer> final answer </answer>
\end{promptbox}

\subsection{Agentic Grounding Tools}
\label{sec:grounding_tools}
The grounding tool $t$ is selected from
$\mathcal{U}=\{\texttt{Prob},\texttt{Cartesian},\texttt{Polar},\texttt{Crop}\}$, whose four interfaces expose complementary outputs from the same grounding backend. Given a referential query, \texttt{Prob} scores the candidate objects in the scene according to how well each matches the query, and returns their
object identities or semantic categories ranked by grounding confidence. \texttt{Cartesian} returns the same ranked candidates together with the 3D center and spatial extent of each object in the global scene coordinate system, providing explicit information about object position and size. \texttt{Polar} transforms each candidate's location into the observer centered coordinate system and reports its horizontal angle relative to the viewing direction and its distance from the observer. \texttt{Crop} retrieves a visual region containing the grounded object and highlights its spatial extent, allowing the agent to inspect appearance information such as color, shape, and local surroundings. These interfaces respectively expose semantic matching, global 3D geometry, viewer relative geometry, and visual appearance. Different questions benefit from different evidence forms: object probabilities may suffice for existence reasoning, while clock direction questions require viewer relative spatial information. CARE-3D therefore treats both the grounding query $g$ and tool selection $t$ as policy decisions, allowing the agent to acquire task appropriate evidence rather than reproducing a prescribed grounding action.

\subsection{Agentic Workflow}
\label{sec:agentic_workflow}
For an input $x$, the agent first samples a structured grounding request $a^{\mathrm{grd}}=(g,t)$, where $g$ is a referential grounding query and $t\in\mathcal{U}$ is the selected grounding tool. Let $\mathcal{G}_{\phi}$ denote the 3D grounding module with parameters $\phi$. The complete interaction is formulated as
\begin{equation}
\begin{aligned}
a^{\mathrm{grd}}=(g,t) &\sim \pi_{\theta}(\cdot\mid x),\\
e &= \mathcal{G}_{\phi}(\mathcal{S},v,g,t),\\
(r,y) &\sim \pi_{\theta}(\cdot\mid x,a^{\mathrm{grd}},e),
\end{aligned}
\label{eq:agentic_workflow}
\end{equation}
where $e$ is the evidence returned by the selected tool, $r$ is the reasoning trace, and $y$ is the generated answer.

Following SceneCOT~\cite{scenecot}, supervised grounding instruction tuning initializes the interaction protocol by teaching the agent to generate executable grounding requests and perform evidence conditioned reasoning, while jointly training the grounding module with annotated object targets. CARE-3D retains this procedure. Reinforcement post-training initializes both components from their supervised checkpoints, freezes the grounding module and vision-language backbone, and updates only the agent's LoRA adapters.

For input $x$, the $i$-th interaction rollout is $\tau_i=(a_i^{\mathrm{grd}},e_i,r_i,y_i)$, where $a_i^{\mathrm{grd}}=(g_i,t_i)$ contains the grounding query $g_i$ and selected tool $t_i\in\mathcal{U}$, $e_i$ is the returned evidence, $r_i$ is the reasoning trace, and $y_i$ is the final answer.

\subsection{Diversified Tool Use in Group Rollouts}
\label{sec:gcs}
Group based reinforcement learning may increasingly concentrate on frequently sampled behaviors, reducing policy entropy and rollout diversity~\cite{dapo,yue2025does}. In CARE-3D, this tendency can cause low frequency grounding tools to disappear from a rollout group before their utility has been adequately evaluated. An absent tool receives no reward based comparison, and CARE cannot evaluate the evidence that it might have returned.

For each training prompt $x$, Group Coverage Sampling (GCS) first samples six free on-policy rollouts, denoted by $\mathcal{T}_{\mathrm{free}}(x)$. Let $\mathcal{T}_{\mathrm{free}}^{\mathrm{valid}}(x)$ denote the subset with valid grounding calls. The coverage count for each tool $u\in\mathcal{U}$ is
\begin{equation}
c_x(u) =
\sum_{\tau_i\in\mathcal{T}_{\mathrm{free}}^{\mathrm{valid}}(x)}
\mathbb{I}[t_i=u].
\label{eq:tool_coverage}
\end{equation}
Malformed calls and unsupported tool identifiers are excluded from the counts. GCS selects the two distinct tools with the lowest counts, including tools with zero occurrences, and generates one coverage rollout for each. The resulting GRPO group contains eight rollouts: six free rollouts and two coverage rollouts.

In each coverage rollout, GCS supplies the tool identifier. The policy generates the grounding query, reasoning trace, and final answer, and the grounding module returns evidence using the supplied tool. The policy loss masks the supplied tool identifier and the returned evidence. The remaining policy-generated tokens are optimized with GRPO. Free and coverage rollouts use the same reward function and participate in the same group-relative advantage computation.

Unlike dynamic sampling in DAPO~\cite{dapo}, which discards groups with uniform outcome rewards, GCS retains all free rollouts and supplements them with underrepresented tools. It changes only rollout group construction, introduces no additional reward, and is disabled during evaluation. Further details are provided in Appendix~\ref{app:gcs_implementation}.

\subsection{Counterfactual Attribution of Returned Evidence}
\label{sec:care}
Final answer rewards do not reveal whether a grounding request returns useful evidence. A correct answer may arise from language priors despite irrelevant evidence, while useful evidence may receive no credit if later reasoning fails. Fixed tool bonuses cannot resolve this ambiguity because they reward tool invocation rather than the utility of its returned evidence. Inspired by likelihood based intermediate credit assignment~\cite{IGPO,infopo}, CARE measures how much the returned evidence supports the ground truth answer.

For rollout $\tau_i$, let $y^\star=(y_1^\star,\ldots,y_L^\star)$ denote the ground truth answer, where $L$ is the number of ground truth answer tokens. Let $a_i^{\mathrm{grd}}$ denote the grounding request and $e_i$ the returned evidence. CARE constructs a counterfactual context by replacing $e_i$ with a fixed evidence masked placeholder $m(e_i)$, while keeping the input and grounding request unchanged.

CARE deliberately excludes the sampled reasoning trace from these scoring contexts. Its purpose is to measure how strongly the returned evidence itself supports the ground truth answer, rather than to evaluate a particular sampled reasoning trajectory. Because reasoning is generated after the evidence is returned, it is an evidence dependent intermediate variable and may already encode the predicted conclusion. Regenerating reasoning under the two contexts would introduce an additional source of variation, whereas reusing the factual reasoning could preserve answer information after the evidence is masked and thereby obscure its contribution. CARE therefore scores the same ground truth answer tokens directly under two matched pre-reasoning contexts that differ only in the availability of returned evidence.

The \emph{Evidence Gain} (EG) is defined as the difference between the mean log likelihoods:
\begin{equation}
\scalebox{0.9}{$\displaystyle
\begin{aligned}
\mathrm{EG}_i = \frac{1}{L}\sum_{k=1}^{L} \Big[ & \log \pi_{\theta} \left( y_k^\star \mid x,a_i^{\mathrm{grd}},e_i \right)
\\
-& \log \pi_{\theta} \left( y_k^\star \mid x,a_i^{\mathrm{grd}},m(e_i)\right) \Big].
\end{aligned}$}
\label{eq:care_delta}
\end{equation}
Here, $\pi_{\theta}$ is the rollout policy snapshot used to generate the current trajectories. It is held fixed during CARE reward computation, and no gradient is propagated through the likelihood evaluation. Both contexts score the same ground truth answer tokens, so their only difference is whether the returned evidence is available. The scoring policy is refreshed between GRPO iterations but remains fixed within each reward computation.

A positive $\mathrm{EG}_i$ indicates that the evidence increases the likelihood of the ground truth answer. A value near zero indicates limited utility, while a negative value suggests distracting or misleading evidence. Because raw likelihood differences may vary with question difficulty and prior confidence, CARE normalizes them among the rollouts generated for the same question. Let $\mu_{\mathrm{EG}}$ and $\sigma_{\mathrm{EG}}$ denote the group mean and standard deviation. The
CARE reward $r_i^{\mathrm{CARE}}$ is
\begin{equation}
\scalebox{1}{$\displaystyle
z_i = \frac{\mathrm{EG}_i-\mu_{\mathrm{EG}}}{\sigma_{\mathrm{EG}}+\epsilon}, \qquad
r_i^{\mathrm{CARE}} = \operatorname{sigmoid}(z_i),$}
\label{eq:care_reward}
\end{equation}
where $z_i$ measures the relative contribution of the evidence returned in rollout $\tau_i$, and $\epsilon$ ensures numerical stability. A positive $z_i$ indicates an above average contribution within the group, whereas a negative value indicates a below average contribution. The sigmoid maps this standardized score to a bounded CARE reward in $(0,1)$, making it compatible with the other reward components.

CARE assigns credit to evidence utility rather than merely to tool use. For example, for a question asking which object is located at the agent's two o'clock, polar evidence receives higher credit when it substantially increases the likelihood of the target object. Evidence whose removal produces little likelihood change receives lower credit. Further details are provided in Appendix~\ref{app:care_implementation}.

\begin{table*}[!t]
    \centering
    \small
    \setlength{\tabcolsep}{4.2pt}
    \renewcommand{\arraystretch}{}
    \resizebox{1\textwidth}{!}{%
    \begin{tabular}{
        lc
        @{\hspace{7pt}}
        rrrrrrr
        @{\hspace{7pt}}
        rr
    }
        \toprule

        \multirow{2}{*}{Method / Backbone}
        & \multirow{2}{*}{Grounded}
        & \multicolumn{7}{c}{MSQA}
        & \multicolumn{2}{c}{Beacon3D} \\

        \cmidrule(lr){3-9}
        \cmidrule(lr){10-11}

        &
        & Count. & Exist. & Attr. & Spatial & Navi. & Others & Overall & Case & Obj. \\

        \midrule
        \multicolumn{11}{l}{\textbf{Baselines}} \\
        \addlinespace[2pt]

        GPT-4o (closed source)
        & $\times$ 
        & 32.3 & 79.3 & \textbf{79.0} & 37.0 & 31.7 & \textbf{91.6} & 52.3 & 57.1 & 20.2 \\

        SplatTalk (Qwen2-7B)
        & $\times$
        & 19.6 & 60.3 & 44.0 & 35.8 & 35.5 & 61.8 & 41.8 & - & - \\

        LEO (Vicuna-7B)
        & $\times$
        & 29.3 & 87.5 & 55.1 & 46.3 & 45.6 & 81.4 & 52.9 & 52.5 & 12.7 \\

        MSR3D (Vicuna-7B)
        & $\times$
        & 32.7 & 87.5 & 53.7 & 44.3 & 51.5 & 72.3 & 52.8 & 51.4 & 11.9 \\

        Chat-Scene (Vicuna-7B-v1.5)
        & $\times$
        & 37.4 & \textbf{92.0} & 49.0 & 47.0 & 58.3 & 83.7 & 56.6 & 53.6 & 14.0 \\

        SceneCOT (Vicuna-7B-v1.5)
        & \tabcheck
        & 47.9 & 82.1 & 49.6 & 47.2 & 51.6 & 80.3 & 55.6 & 58.9 & 23.2 \\

        \midrule
        \multicolumn{11}{l}{
            \textbf{Ours}
        } \\
        \addlinespace[2pt]
        
        SFT (LLaVA-v1.5-7B)
        & \tabcheck
        & 49.2 & 87.5 & 54.4 & 47.9 & 44.9 & 83.4 & 56.5 & 59.1 & 23.1 \\
        \rowcolor{blue!7}
        \textbf{CARE-3D} (LLaVA-v1.5-7B)
        & \tabcheck
        & 54.9 & \underline{88.7} & 61.9 & 58.6 & 57.6 & 84.1 & 64.2 & 64.0 & \underline{26.1} \\
        $\Delta$ & & {\dpos{5.7}} & {\dpos{1.2}} & {\dpos{7.5}} & {\dpos{10.7}} & {\dpos{12.7}} & {\dpos{0.7}} & {\dpos{7.7}} & {\dpos{4.9}} & {\dpos{3.0}} \\
        \midrule

        SFT (Qwen2.5-VL-3B)
        & \tabcheck
        & 45.1 & 62.5 & 54.1 & 53.5 & 56.4 & 84.6 & 56.6 & 57.7 & 19.7 \\

        \rowcolor{blue!7}
        \textbf{CARE-3D} (Qwen2.5-VL-3B)
        & \tabcheck
        & {51.2} & 77.1 & \underline{67.4} & \underline{59.7} & {60.9} & 88.2 & \underline{64.5}
        & {61.8} & {23.3} \\
        $\Delta$ & & {\dpos{6.1}} & {\dpos{14.6}} & {\dpos{13.3}} & {\dpos{6.2}} & {\dpos{4.5}} & {\dpos{3.6}} & {\dpos{7.9}} & {\dpos{4.1}} & {\dpos{3.6}} \\
        \midrule

        SFT (Gemma-3-4B-IT)
        & \tabcheck
        & 53.8 & 81.2 & 48.8 & 56.2 & 58.7 & 78.4 & 59.9 & 60.4 & 23.4 \\

        \rowcolor{blue!7}
        \textbf{CARE-3D} (Gemma-3-4B-IT)
        & \tabcheck
        & \underline{55.6} & 84.2 & 55.5 & 57.2 & \textbf{63.1} & 88.4 & {63.6} & \textbf{65.6} & \textbf{26.5} \\
        $\Delta$ & & {\dpos{1.8}} & {\dpos{3.0}} & {\dpos{6.7}} & {\dpos{1.0}} & {\dpos{4.4}} & {\dpos{10.0}} & {\dpos{3.7}} & {\dpos{5.2}} & {\dpos{3.1}} \\

        \midrule

        SFT (Qwen3-VL-4B)
        & \tabcheck
        & 49.8 & 72.9 & 55.8 & 51.4 & 56.2 & 83.8 & 58.1 & 59.6 & 21.2 \\

        \rowcolor{blue!7}
        \textbf{CARE-3D} (Qwen3-VL-4B)
        & \tabcheck
        & \textbf{56.6} & 79.0 & 63.3 & \textbf{62.2} & \underline{61.9} & \underline{88.5} & \textbf{65.7}
        & \underline{64.5} & {25.7} \\
        $\Delta$ & & {\dpos{6.8}} & {\dpos{6.1}} & {\dpos{7.5}} & {\dpos{10.8}} & {\dpos{5.7}} & {\dpos{4.7}} & {\dpos{7.6}} & {\dpos{4.9}} & {\dpos{4.5}} \\
        \bottomrule
    \end{tabular}%
    }
    \caption{Comparison with existing methods and different backbone models. SFT denotes the corresponding supervised baseline for each backbone. All scores are reported on a 0-100 scale. \textbf{Bold} and \underline{underlined} values denote the best and second best results. Grounded indicates whether the model produces explicit object linked grounding evidence before answering. LLaVA-v1.5-7B shares the same Vicuna-7B-v1.5 language backbone as the baseline methods, enabling a same backbone comparison.}
    \label{tab:sota_comparison}
\end{table*}

\subsection{Overall Reward and Policy Update}
\label{sec:overall_reward}

Each rollout receives a scalar reward combining interaction validity, answer correctness, and returned evidence utility:
\begin{equation}
\scalebox{1}{$\displaystyle
R_i = w_{\mathrm{fmt}}r_i^{\mathrm{fmt}} + w_{\mathrm{ans}}r_i^{\mathrm{ans}} + w_{\mathrm{CARE}}r_i^{\mathrm{CARE}} - p_i,
$}
\label{eq:overall_reward}
\end{equation}
where $r_i^{\mathrm{fmt}}$ evaluates whether the rollout follows the required interaction format, $r_i^{\mathrm{ans}}$ evaluates the semantic correctness of the generated answer, $r_i^{\mathrm{CARE}}$ evaluates the utility of the returned evidence, and $p_i$ denotes the invalid interaction for each rollout. The answer reward maps the judge's score from \(\{1,\ldots,5\}\) linearly to \([0,1]\). See Box~2 in Appendix~\ref{app:prompts} for the judge prompt and Appendix~\ref{app:reward_config} for the reward configuration. The coefficients $w_{\mathrm{fmt}}$, $w_{\mathrm{ans}}$, and $w_{\mathrm{CARE}}$ control the corresponding reward contributions.

The penalty $p_i$ is zero for a valid interaction and positive when the grounding request is malformed, specifies an unsupported tool, or cannot be executed. Free and coverage rollouts use the same reward function. GCS modifies only the rollout group, whereas CARE contributes directly to the scalar reward. The resulting $R_i$ is used in Equation~\ref{eq:grpo_advantage}, and the agent is optimized with Equation~\ref{eq:grpo_objective}. Only the agent's LoRA adapters are updated; the vision-language backbone and 3D grounding module remain frozen.

\begin{table*}[h]
    \centering
    \setlength{\tabcolsep}{4.2 pt}
    \renewcommand{\arraystretch}{1}
    \resizebox{\textwidth}{!}{%
    \begin{tabular}{
        llcc
        @{\hspace{7pt}}
        ccccccc
        @{\hspace{7pt}}
        cc
    }
        \toprule
        \multirow{2}{*}{Training} & \multirow{2}{*}{Model / Variant} & \multirow{2}{*}{CARE} & \multirow{2}{*}{GCS}
        & \multicolumn{7}{c}{MSQA} & \multicolumn{2}{c}{Beacon3D} \\
        \cmidrule(lr){5-11}
        \cmidrule(lr){12-13}
        & & & &
        Count. & Exist. & Attr./Desc. & Spat./Refer. & Nav. & Other & Overall & Case & Obj. \\
        \midrule
        SFT & Qwen3-VL-4B & - & - & 49.8 & 72.9 & 55.8 & 51.4 & 56.2 & 83.8 & 58.1 & 59.6 & 21.2 \\
        \midrule
        \multirow{4}{*}{GRPO} & Answer + Format & - & -
        & \vc{53.4}{\dpos{3.6}}
        & \vc{76.0}{\dpos{3.1}}
        & \vc{62.1}{\dpos{6.3}}
        & \vc{56.2}{\dpos{4.8}}
        & \vc{60.2}{\dpos{4.0}}
        & \vc{\tabbest{88.5}}{\dpos{4.7}}
        & \vc{62.6}{\dpos{4.5}}
        & \vc{61.6}{\dpos{2.0}}
        & \vc{22.9}{\dpos{1.7}} \\
        
        & + CARE
        & \tabcheck & -
        & \vc{54.7}{\dpos{4.9}}
        & \vc{71.9}{\dneg{1.0}}
        & \vc{\tabsecond{62.8}}{\dpos{7.0}}
        & \vc{60.2}{\dpos{8.8}}
        & \vc{58.3}{\dpos{2.1}}
        & \vc{87.8}{\dpos{4.0}}
        & \vc{63.2}{\dpos{5.1}}
        & \vc{\tabsecond{62.6}}{\dpos{3.0}}
        & \vc{\tabsecond{25.4}}{\dpos{4.2}} \\
        
        & + GCS
        & - & \tabcheck
        & \vc{\tabsecond{56.2}}{\dpos{6.4}}
        & \vc{\tabsecond{78.2}}{\dpos{5.3}}
        & \vc{60.3}{\dpos{4.5}}
        & \vc{\tabsecond{61.3}}{\dpos{9.9}}
        & \vc{\tabsecond{61.6}}{\dpos{5.4}}
        & \vc{\tabsecond{88.2}}{\dpos{4.4}}
        & \vc{\tabsecond{64.7}}{\dpos{6.6}}
        & \vc{62.4}{\dpos{2.8}}
        & \vc{23.7}{\dpos{2.5}} \\
        \rowcolor{blue!7}
        
        & \textbf{CARE-3D}
        & \tabcheck & \tabcheck
        & \vc{\tabbest{56.6}}{\dpos{6.8}}
        & \vc{\tabbest{79.0}}{\dpos{6.1}}
        & \vc{\tabbest{63.3}}{\dpos{7.5}}
        & \vc{\tabbest{62.2}}{\dpos{10.8}}
        & \vc{\tabbest{61.9}}{\dpos{5.7}}
        & \vc{\tabbest{88.5}}{\dpos{4.7}}
        & \vc{\tabbest{65.7}}{\dpos{7.6}}
        & \vc{\tabbest{64.5}}{\dpos{4.9}}
        & \vc{\tabbest{25.7}}{\dpos{4.5}} \\
        \bottomrule
    \end{tabular}%
    }
    \caption{
        Ablation study of \textsc{CARE-3D} on MSQA and Beacon3D. CARE and GCS denote Counterfactual Attribution of Returned Evidence and Group Coverage Sampling, respectively.
        \textbf{Bold} and \underline{underlined} indicate the best and second best results among GRPO variants; small numbers in parentheses give the improvement over the SFT model.
    }
    \label{tab:main_ablation}
\end{table*}

\subsection{Reinforcement Post-Training with GRPO}
Supervised tuning imitates one grounding request per question without comparing alternative tools. CARE-3D therefore applies GRPO~\cite{DeepSeek,DeepSeekMath} to optimize complete grounded interactions.

For input $x$, the policy constructs $G$ rollouts
$\mathcal{T}(x)=\{\tau_i\}_{i=1}^{G}$ with scalar rewards
$\{R_i\}_{i=1}^{G}$. The group relative advantage is
\begin{equation}
A_i=\frac{R_i-\mu_R}{\sigma_R+\delta},
\label{eq:grpo_advantage}
\end{equation}
where $\mu_R$ and $\sigma_R$ are the mean and standard deviation of the group rewards, and $\delta$ ensures numerical stability.

Let $w_{i,k}$ denote the $k$-th optimized token in $\tau_i$ and
$h_{i,k}$ its preceding context. The policy ratio is
\begin{equation}
\rho_{i,k}(\theta)
=
\frac{\pi_{\theta}(w_{i,k}\mid h_{i,k})}
{\pi_{\theta_{\mathrm{old}}}(w_{i,k}\mid h_{i,k})},
\label{eq:policy_ratio}
\end{equation}
where $\pi_{\theta_{\mathrm{old}}}$ is the rollout policy. Defining
$\bar{\rho}_{i,k}=\operatorname{clip}
(\rho_{i,k},1-\varepsilon,1+\varepsilon)$, GRPO optimizes
\begin{equation}
\scalebox{0.85}{$\displaystyle
\begin{aligned}
\mathcal{J}_{\mathrm{GRPO}}(\theta)
&= \mathbb{E}_{x,\mathcal{T}(x)} \Bigg[
\frac{1}{G}\sum_{i=1}^{G} \frac{1}{K_i}\sum_{k=1}^{K_i} \\
&\qquad \Big( \min\{\rho_{i,k}A_i,\ \bar{\rho}_{i,k}A_i\}
- \beta D_{i,k}^{\mathrm{KL}} \Big) \Bigg].
\end{aligned}
$}
\label{eq:grpo_objective}
\end{equation}
Here, $K_i$ is the number of optimized tokens, $\varepsilon$ is the clipping threshold, and $\beta$ controls regularization against a fixed reference policy. The two averages normalize over tokens and rollouts.

\section{Experiment}

\subsection{Setup}
\paragraph{Datasets.} CARE-3D is evaluated on MSQA and Beacon3D~\cite{msqa,scenecot,Beacon3D}. Supervised traces from MSQA and GQA3D are used for SFT, whereas reinforcement post-training is conducted only on the MSQA training split. Beacon3D is reserved exclusively for evaluation.

\paragraph{Baselines.}
CARE-3D is compared with GPT-4o~\cite{gpt-4o}, SplatTalk~\cite{Splattalk}, LEO~\cite{LEO}, MSR3D~\cite{msqa}, Chat-Scene~\cite{ChatScene}, and SceneCOT~\cite{scenecot}. The corresponding SFT policy and an \emph{Answer+Format} GRPO baseline are additionally included to isolate the effects of reinforcement post-training and the proposed components.

\paragraph{Evaluation Metrics.}
Following official protocols, GPT-Score is used to evaluate semantic correctness. MSQA reports six category scores and an overall score, while Beacon3D reports per question \emph{Case} and the stricter object level \emph{Obj.}, which requires all questions for a target object to be answered correctly. Scores range from 0 to 100. The training time reward judge is not used for final evaluation.

\paragraph{Detailed implementation} and training details are provided in Appendix~\ref{app:experimental_setup} and~\ref{app:training_configurations}, respectively.

\subsection{Overall Performance}

Table~\ref{tab:sota_comparison} compares CARE-3D with representative 3D vision-language methods across four backbones: Qwen3-VL-4B-Instruct~\cite{Qwen3-VL}, Qwen2.5-VL-3B-Instruct~\cite{Qwen2.5-VL}, LLaVA-v1.5-7B~\cite{llava}, and Gemma-3-4B-IT~\cite{gemma3}. With Qwen3-VL-4B-Instruct, CARE-3D achieves the highest overall MSQA score and outperforms SceneCOT in five of six categories, with the largest gain in spatial/referential reasoning.

Across all four backbones, CARE-3D improves the MSQA overall score and both Beacon3D metrics over the corresponding SFT policies. These consistent gains indicate that CARE-3D is effective across different vision encoders, language backbones, and multimodal alignment strategies.

\begin{figure}[h]
    \centering
    \includegraphics[width=\columnwidth]
    {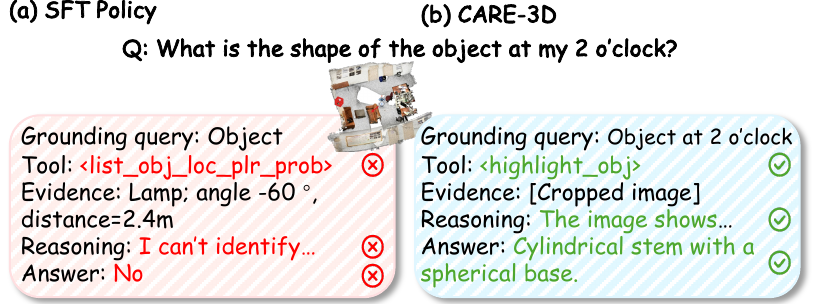}
    \caption{Qualitative comparison on an attribute question. CARE-3D selects highlighted visual evidence and correctly recognizes the target object's shape. }
    \label{fig:case_attribute}
\end{figure}

Figure~\ref{fig:case_attribute} compares the SFT policy with CARE-3D on the same attribute question. The SFT policy retrieves location evidence that identifies the target but does not reveal its shape, resulting in an incorrect answer. CARE-3D instead selects highlighted visual evidence and correctly identifies the target object's shape, demonstrating \textbf{better alignment between evidence acquisition and the information requested by the question}. More examples are provided in Appendix~\ref{app:qualitative}.

\subsection{Ablation Study}
Table~\ref{tab:main_ablation} examines the contributions of CARE and GCS. The component variants exhibit complementary effects. GCS provides stronger gains on MSQA by preserving grounding tool coverage, whereas CARE yields clearer improvements on the object centric Beacon3D metrics through evidence sensitive credit assignment. Combining both achieves the strongest performance, suggesting that tool exploration and evidence attribution address distinct aspects of policy optimization.

Compared with the SFT policy, the \emph{Answer+Format} baseline improves overall performance, confirming the general benefit of reinforcement post-training. CARE further strengthens object level consistency on Beacon3D, whereas GCS produces broader gains across MSQA reasoning categories. Combining both components achieves the best results on both benchmarks, demonstrating improvements beyond answer and format only GRPO.

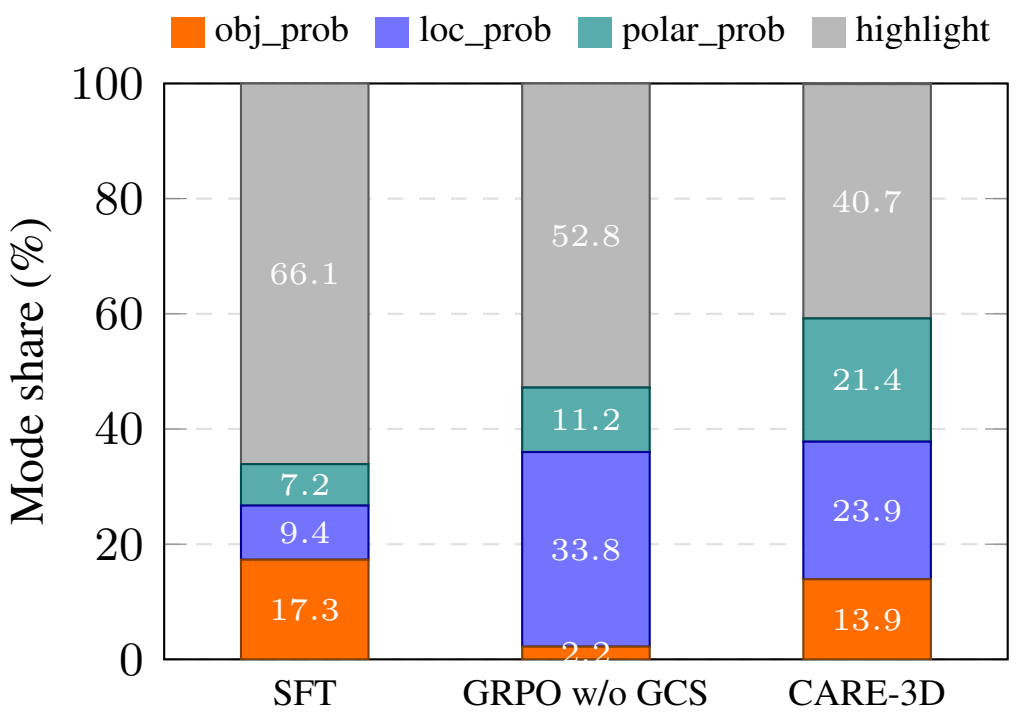
\begin{figure}[!t]
\centering
\begin{tikzpicture}[scale=1, every node/.style={transform shape}]
\begin{axis}[
    ybar stacked,
    bar width=24pt,
    width=0.93\columnwidth,
    height=0.7\columnwidth,
    ymin=0, ymax=100,
    ylabel={Mode share (\%)},
    ylabel style={font=\footnotesize, yshift=-4pt},
    symbolic x coords={SFT, {GRPO w/o GCS}, {CARE-3D}},
    xtick=data,
    x tick label style={font=\scriptsize},
    tick label style={font=\footnotesize},
    enlarge x limits=0.25,
    legend style={
        font=\scriptsize,
        at={(0.5,1.02)}, anchor=south,
        legend columns=4,
        draw=none, fill=none,
        /tikz/every even column/.append style={column sep=3pt},
    },
    legend image code/.code={
        \draw[#1, draw=none] (0cm,-0.1cm) rectangle (0.22cm,0.1cm);
    },
    ymajorgrids, grid style={dashed, gray!25},
    nodes near coords,
    nodes near coords style={font=\tiny, color=white, anchor=center},
    every node near coord/.append style={/pgf/number format/fixed,
        /pgf/number format/precision=1},
]
\addplot[fill=orange!85!red, draw=orange!50!black]
    coordinates {(SFT,17.3) ({GRPO w/o GCS},2.2) ({CARE-3D},13.9)};
\addplot[fill=blue!55!white, draw=blue!60!black]
    coordinates {(SFT,9.4) ({GRPO w/o GCS},33.8) ({CARE-3D},23.9)};
\addplot[fill=teal!65!white, draw=teal!60!black]
    coordinates {(SFT,7.2) ({GRPO w/o GCS},11.2) ({CARE-3D},21.4)};
\addplot[fill=gray!55!white, draw=gray!70!black]
    coordinates {(SFT,66.1) ({GRPO w/o GCS},52.8) ({CARE-3D},40.7)};
\legend{obj\_prob, loc\_prob, polar\_prob, highlight}
\end{axis}
\end{tikzpicture}
\caption{Distribution of grounding-tool selections on MSQA at inference time. Each segment shows the percentage of questions for which the SFT policy, GRPO without GCS, or CARE-3D selects the corresponding grounding tool. GCS is disabled during inference.
}
\label{fig:mode_distribution}
\end{figure}

GCS supplements free rollouts with loss masked coverage rollouts from underrepresented grounding tools. To evaluate its contribution, the \emph{Answer+Format} baseline is compared with and without GCS. Adding GCS improves the MSQA overall score from 62.6 to 64.7, while removing it from CARE-3D reduces performance on both MSQA and Beacon3D, showing the benefit of maintaining diverse rollout groups.

\subsection{Effectiveness of GCS}

Figure~\ref{fig:mode_distribution} shows how GCS during training
affects tool selection at inference. The policy trained with GCS
selects \texttt{obj\_prob} and \texttt{polar\_prob} more frequently
than GRPO without GCS, yielding a more balanced distribution across
the four grounding tools. At inference, the trained policy selects
the grounding tool directly.

\begin{table}[t]
    \centering
    \footnotesize
    \setlength{\tabcolsep}{4pt}
    \begin{tabular}{lccc}
        \toprule
        \multirow{2}{*}{$K$}
        & MSQA
        & \multicolumn{2}{c}{Beacon3D} \\
        \cmidrule(lr){2-2}
        \cmidrule(lr){3-4}
        & Overall & Case & Obj. \\
        \midrule
        0          & 63.2 & 62.6 & 25.4 \\
        1          & 64.5 & 63.1 & 25.1 \\
        2 (random) & 61.7 & 61.3 & 22.8 \\
        2 (GCS)    & \textbf{65.7} & \textbf{64.5} & \textbf{25.7} \\
        4          & 64.4 & 60.9 & 23.4 \\
        \bottomrule
    \end{tabular}
    \caption{Effect of the number of coverage rollouts $K$ and the
    tool selection rule. Each group contains eight rollouts.
    The random variant selects coverage tools at random. All other
    rows with $K>0$ select the least-covered tools.}
    \label{tab:gcs_sensitivity}
\end{table}

Table~\ref{tab:gcs_sensitivity} varies the number of coverage
rollouts $K$ while keeping the total group size fixed at eight.
GCS with $K=2$ achieves the highest scores on all three metrics.
At $K=4$, the group contains four free rollouts and four coverage
rollouts, and performance decreases on both benchmarks.

The two $K=2$ variants use the same budget of six free and two
coverage rollouts per prompt. Selecting the least-covered tools
improves MSQA Overall by 4.0 points and Beacon3D Case and Obj.\
by 3.2 and 2.9 points, respectively, over random tool selection.

\subsection{Effectiveness of CARE}
\begin{table}[h]
    \centering
    \footnotesize
    \setlength{\tabcolsep}{4pt}
    \renewcommand{\arraystretch}{}

    \begin{tabular}{lccc}
        \toprule
        \multirow{2}{*}{Counterfactual}
        & \multicolumn{1}{c}{MSQA}
        & \multicolumn{2}{c}{Beacon3D} \\
        \cmidrule(lr){2-2}
        \cmidrule(lr){3-4}
        & Overall & Case & Obj. \\
        \midrule
        Random Tokens    & 62.2 & 56.5 & 13.9 \\
        Alternative Tool & 21.2 & 14.5 & 2.1 \\
        Evidence Masking (CARE-3D) & \textbf{65.7} & \textbf{64.5} & \textbf{25.7} \\
        \bottomrule
    \end{tabular}
    \caption{Ablation of the counterfactual construction used by CARE-3D. All variants use the same initialization and training configuration, differing only in the counterfactual construction.}
    \label{tab:care_counterfactual}
\end{table}

CARE assigns evidence sensitive credit by measuring how returned evidence changes the likelihood of the ground truth answer. As shown in Table~\ref{tab:main_ablation}, CARE improves both MSQA and Beacon3D, particularly the object centric Beacon3D metrics. Table~\ref{tab:care_counterfactual} further compares random token, alternative tool, and evidence masking counterfactuals. Evidence masking performs best because it preserves the question and grounding action while removing only the returned evidence.

Among 317 SFT errors on MSQA, 103 predictions remain unchanged after evidence masking (\emph{answer invariant errors}). We evaluate CARE-3D on this fixed subset.
\begin{table}[t]
    \centering
    \footnotesize
    \setlength{\tabcolsep}{5pt}
    \renewcommand{\arraystretch}{1}
\begin{tabular}{lccc}
    \toprule
    QA Type & $N$ & Recovered & $\overline{D}_{\mathrm{KL}}$ \\
    \midrule
    Counting        & 11  & 7 (63.6\%)  & 0.57 \\
    Existence       & 6   & 4 (66.7\%)  & 0.22 \\
    Attr./Desc.     & 58  & 26 (44.8\%) & 0.68 \\
    Spatial/Refer.  & 23  & 7 (30.4\%)  & 0.07 \\
    Navigation      & 4   & 2 (50.0\%)  & 0.15 \\
    Others          & 1   & 1 (100\%)   & 0.11 \\
    \midrule
    Overall         & 103 & 47 (45.6\%) & 0.48 \\
    \bottomrule
\end{tabular}
\caption{
        Recovery and evidence sensitivity of CARE-3D on answer invariant errors made by the SFT policy.
        $\overline{D}_{\mathrm{KL}}$ denotes the mean per question Full Masked token level divergence.
    }
    \label{tab:care_error_recovery}
\end{table}
As shown in Table~\ref{tab:care_error_recovery}, CARE-3D corrects 47 of these 103 failures, corresponding to a 45.6\% recovery rate. Its Full Masked divergence also shows measurable sensitivity to evidence interventions, particularly for attribute/description and counting questions. Together, these results show that \textbf{CARE reduces evidence insensitive errors and strengthens reliance on grounded evidence}. Further analyses are provided in Appendix~\ref{app:evidence_analysis}.

\paragraph{Evidence interventions.}
We keep the question fixed and modify either relevant or irrelevant returned evidence. Relevant modifications change the correct answer, whereas irrelevant modifications preserve it. Predictions are evaluated against the corresponding answers after each intervention.

\begin{table}[h]
\centering
\small
\setlength{\tabcolsep}{4pt}
\begin{tabular}{lccc}
\hline
Policy &
\shortstack{Original\\Acc.$\uparrow$} &
\shortstack{Relevant edits\\Acc.$\uparrow$} &
\shortstack{Irrelevant edits\\Acc.$\uparrow$} \\
\hline
SFT  & 49.7\% & 43.4\% ($-6.3$) & 49.5\% ($-0.2$) \\
CARE-3D & 56.8\% & 55.9\% ($-0.9$) & 56.8\% ($0.0$) \\
\hline
\end{tabular}
\caption{Answer accuracy under evidence interventions. Values in parentheses indicate percentage-point changes relative to the original evidence condition.}
\label{tab:evidence_interventions}
\end{table}

As shown in Table~\ref{tab:evidence_interventions}, relevant modifications reduce SFT accuracy by 6.3 percentage points compared with only 0.9 percentage points for CARE-3D. Under irrelevant modifications, CARE-3D maintains its original accuracy, while SFT decreases by 0.2 percentage points. These results support better performance under answer-changing relevant interventions and robustness to answer-preserving irrelevant interventions.

\section{Conclusion}
CARE-3D is a reinforcement post-training framework for evidence grounded situated 3D reasoning. CARE assigns credit to returned evidence, whereas GCS preserves tool coverage during optimization. Across MSQA and Beacon3D, their combination outperforms SFT and answer and format GRPO, particularly in spatial reasoning and grounding answer coherence. These findings highlight the importance of jointly optimizing evidence acquisition and use for reliable situated 3D reasoning.

\section*{Limitations}        
CARE-3D currently performs a single grounding action per question and therefore does not support iterative query refinement or multi tool composition. Future work could extend the framework to multi step grounding, allowing the agent to refine unsuccessful requests and combine complementary evidence.

CARE estimates evidence utility by masking returned evidence and scoring the ground truth answer. This counterfactual may introduce distribution shift, while likelihood scores can be affected by answer phrasing. Future work may explore in distribution counterfactuals, semantically normalized scoring, and reasoning aware credit assignment. CARE also increases training time computation by evaluating factual and evidence masked contexts, although it introduces no additional inference time cost.

\paragraph{Potential risks.}
Although CARE-3D is designed for research, incorrect grounding or spatial reasoning could lead to unsafe behavior if deployed in embodied systems. Its outputs should therefore not be used in safety-critical settings without human oversight and additional validation.

\section*{Acknowledgments}
This research was supported in part by the Australian Research Council (ARC) under grants IC230100016, CE200100025, DP230101196, and DE250100919.

\bibliography{custom}   

@inproceedings{scenecot,
  title={SceneCOT: Eliciting Grounded Chain-of-Thought Reasoning in 3D Scenes},
  author={Linghu, Xiongkun and 
  Huang, Jiangyong and 
  Zhu, Ziyu and 
  Jia, Baoxiong and 
  Huang, Siyuan},
  booktitle={ICLR},
  year={2026}
}

@inproceedings{Beacon3D,
  title={Unveiling the Mist over 3D Vision-Language Understanding: Object-centric Evaluation with Chain-of-Analysis},
  author={Huang, Jiangyong and 
  Jia, Baoxiong and 
  Wang, Yan and 
  Zhu, Ziyu and 
  Linghu, Xiongkun and 
  Li, Qing and Zhu, 
  Song-Chun and 
  Huang, Siyuan},
  booktitle={CVPR},
  year={2025}
}

@article{taco,
  author       = {Mingkuan Feng and
                  Jinyang Wu and
                  Hao Gu and
                  Fangrui Lv and
                  Ruihan Jin and
                  ChuYuan Zhang and
                  Zhengqi Wen and
                  Jianhua Tao},
  title        = {{TACO:} Tool-Augmented Credit Optimization for Agentic Tool Use},
  journal      = {CoRR},
  volume       = {abs/2606.30251},
  year         = {2026},
}

@article{DeepSeek,
   title={DeepSeek-R1 incentivizes reasoning in LLMs through reinforcement learning},
   journal={Nature},
   author={Guo, Daya and Yang, Dejian and Zhang, Haowei and Song, Junxiao and Wang, Peiyi and Zhu, Qihao and Xu, Runxin and Zhang, Ruoyu and Ma, Shirong and Bi, Xiao and Zhang, Xiaokang and Yu, Xingkai and Wu, Yu and Wu, Z. F. and Gou, Zhibin and Shao, Zhihong and Li, Zhuoshu and Gao, Ziyi and Liu, Aixin and Xue, Bing and Wang, Bingxuan and Wu, Bochao and Feng, Bei and Lu, Chengda and Zhao, Chenggang and Deng, Chengqi and Ruan, Chong and Dai, Damai and Chen, Deli and Ji, Dongjie and Li, Erhang and Lin, Fangyun and Dai, Fucong and Luo, Fuli and Hao, Guangbo and Chen, Guanting and Li, Guowei and Zhang, H. and Xu, Hanwei and Ding, Honghui and Gao, Huazuo and Qu, Hui and Li, Hui and Guo, Jianzhong and Li, Jiashi and Chen, Jingchang and Yuan, Jingyang and Tu, Jinhao and Qiu, Junjie and Li, Junlong and Cai, J. L. and Ni, Jiaqi and Liang, Jian and Chen, Jin and Dong, Kai and Hu, Kai and You, Kaichao and Gao, Kaige and Guan, Kang and Huang, Kexin and Yu, Kuai and Wang, Lean and Zhang, Lecong and Zhao, Liang and Wang, Litong and Zhang, Liyue and Xu, Lei and Xia, Leyi and Zhang, Mingchuan and Zhang, Minghua and Tang, Minghui and Zhou, Mingxu and Li, Meng and Wang, Miaojun and Li, Mingming and Tian, Ning and Huang, Panpan and Zhang, Peng and Wang, Qiancheng and Chen, Qinyu and Du, Qiushi and Ge, Ruiqi and Zhang, Ruisong and Pan, Ruizhe and Wang, Runji and Chen, R. J. and Jin, R. L. and Chen, Ruyi and Lu, Shanghao and Zhou, Shangyan and Chen, Shanhuang and Ye, Shengfeng and Wang, Shiyu and Yu, Shuiping and Zhou, Shunfeng and Pan, Shuting and Li, S. S. and Zhou, Shuang and Wu, Shaoqing and Yun, Tao and Pei, Tian and Sun, Tianyu and Wang, T. and Zeng, Wangding and Liu, Wen and Liang, Wenfeng and Gao, Wenjun and Yu, Wenqin and Zhang, Wentao and Xiao, W. L. and An, Wei and Liu, Xiaodong and Wang, Xiaohan and Chen, Xiaokang and Nie, Xiaotao and Cheng, Xin and Liu, Xin and Xie, Xin and Liu, Xingchao and Yang, Xinyu and Li, Xinyuan and Su, Xuecheng and Lin, Xuheng and Li, X. Q. and Jin, Xiangyue and Shen, Xiaojin and Chen, Xiaosha and Sun, Xiaowen and Wang, Xiaoxiang and Song, Xinnan and Zhou, Xinyi and Wang, Xianzu and Shan, Xinxia and Li, Y. K. and Wang, Y. Q. and Wei, Y. X. and Zhang, Yang and Xu, Yanhong and Li, Yao and Zhao, Yao and Sun, Yaofeng and Wang, Yaohui and Yu, Yi and Zhang, Yichao and Shi, Yifan and Xiong, Yiliang and He, Ying and Piao, Yishi and Wang, Yisong and Tan, Yixuan and Ma, Yiyang and Liu, Yiyuan and Guo, Yongqiang and Ou, Yuan and Wang, Yuduan and Gong, Yue and Zou, Yuheng and He, Yujia and Xiong, Yunfan and Luo, Yuxiang and You, Yuxiang and Liu, Yuxuan and Zhou, Yuyang and Zhu, Y. X. and Huang, Yanping and Li, Yaohui and Zheng, Yi and Zhu, Yuchen and Ma, Yunxian and Tang, Ying and Zha, Yukun and Yan, Yuting and Ren, Z. Z. and Ren, Zehui and Sha, Zhangli and Fu, Zhe and Xu, Zhean and Xie, Zhenda and Zhang, Zhengyan and Hao, Zhewen and Ma, Zhicheng and Yan, Zhigang and Wu, Zhiyu and Gu, Zihui and Zhu, Zijia and Liu, Zijun and Li, Zilin and Xie, Ziwei and Song, Ziyang and Pan, Zizheng and Huang, Zhen and Xu, Zhipeng and Zhang, Zhongyu and Zhang, Zhen},
   year={2025},
   }

@inproceedings{VISPROG,
  author       = {Tanmay Gupta and
                  Aniruddha Kembhavi},
  title        = {Visual Programming: Compositional visual reasoning without training},
  booktitle    = {CVPR},
  year         = {2023},
}

@inproceedings{3dllm,
 author = {Hong, Yining and 
    Zhen, Haoyu and 
    Chen, Peihao and 
    Zheng, Shuhong and 
    Du, Yilun and 
    Chen, Zhenfang and 
    Gan, Chuang},
 title = {3D-LLM: Injecting the 3D World into Large Language Models},
 journal = {NeurIPS},
 year = {2023},
}

@inproceedings{pointllm,
  title={PointLLM: Empowering Large Language Models to Understand Point Clouds},
  author={Xu, Runsen and 
    Wang, Xiaolong and 
    Wang, Tai and 
    Chen, Yilun and 
    Pang, Jiangmiao and 
    Lin, Dahua},
  booktitle={ECCV},
  year={2024}
}

@inproceedings{LEO,
  title={An Embodied Generalist Agent in 3D World},
  author={Huang, Jiangyong and 
      Yong, Silong and 
      Ma, Xiaojian and 
      Linghu, Xiongkun and 
      Li, Puhao and 
      Wang, Yan and 
      Li, Qing and 
      Zhu, Song-Chun and 
      Jia, Baoxiong and 
      Huang, Siyuan},
  booktitle={ICML},
  year={2024}
}

@article{grounded3dllmreferenttokens,
  author       = {Yilun Chen and
                  Shuai Yang and
                  Haifeng Huang and
                  Tai Wang and
                  Ruiyuan Lyu and
                  Runsen Xu and
                  Dahua Lin and
                  Jiangmiao Pang},
  title        = {Grounded 3D-LLM with Referent Tokens},
  journal      = {CoRR},
  volume       = {abs/2405.10370},
  year         = {2024},
}

@InProceedings{llava3d,
    author = {Zhu, Chenming and Wang, Tai and Zhang, Wenwei and Pang, Jiangmiao and Liu, Xihui},
    title = {LLaVA-3D: A Simple yet Effective Pathway to Empowering LMMs with 3D Capabilities},
    booktitle = {Proceedings of the IEEE/CVF International Conference on Computer Vision (ICCV)},
    year = {2025},
}

@inproceedings{msqa,
  title={Multi-modal Situated Reasoning in 3D Scenes},
  author={Linghu, Xiongkun and Huang, Jiangyong and Niu, Xuesong and Ma, Xiaojian and Jia, Baoxiong and Huang, Siyuan},
  booktitle = {Advances in Neural Information Processing Systems},
  year={2024}
}

@article{RieMind,
  author       = {Fernando Ropero and
                  Erkin Turkoz and
                  Daniel Matos and
                  Junqing Du and
                  Antonio Ruiz and
                  Yanfeng Zhang and
                  Lu Liu and
                  Mingwei Sun and
                  Yongliang Wang},
  title        = {RieMind: Geometry-Grounded Spatial Agent for Scene Understanding},
  journal      = {CoRR},
  volume       = {abs/2603.15386},
  year         = {2026},
}

@article{think3d,
  author       = {Zaibin Zhang and
                  Yuhan Wu and
                  Lianjie Jia and
                  Yifan Wang and
                  Zhongbo Zhang and
                  Yijiang Li and
                  Binghao Ran and
                  Fuxi Zhang and
                  Zhuohan Sun and
                  Zhenfei Yin and
                  Lijun Wang and
                  Huchuan Lu},
  title        = {Think3D: Thinking with Space for Spatial Reasoning},
  journal      = {CoRR},
  volume       = {abs/2601.13029},
  year         = {2026},

}

@inproceedings{skill3d,
    title = "{S}kill-3{D}: Evolving Scene-Aware Skills for Agentic 3{D} Spatial Reasoning",
    author = "Li, Haoyuan  and
      Hu, Zhengdong  and
      Wang, Jun  and
      Fan, Hehe  and
      Yang, Yi",
    booktitle = "Findings of the {A}ssociation for {C}omputational {L}inguistics: {EMNLP} 2026",
    year = "2026",

}

@inproceedings{deepeyes,
 author = {Zheng, Ziwei and Yang, Minghao and Hong, Jack and Zhao, Chenxiao and Xu, Guohai and Yang, Le and Shen, Chao and XingYu},
 booktitle = {International Conference on Learning Representations},
 title = {DeepEyes: Incentivizing "Thinking with Images" via Reinforcement Learning},
 year = {2026}
}

@inproceedings{VTool-R1,
 author = {Wu, Mingyuan and Yang, Jingcheng and Jiang, Jize and Li, Meitang and Yan, Kaizhuo and Yu, Hanchao and Zhang, Minjia and Zhai, ChengXiang and Nahrstedt, Klara},
 booktitle = {International Conference on Learning Representations},
 title = {VTool-R1: VLMs Learn to Think with Images via Reinforcement Learning on Multimodal Tool Use},
 year = {2026}
}

@inproceedings{dapo,
  author       = {Qiying Yu and
                  Zheng Zhang and
                  Ruofei Zhu and
                  Yufeng Yuan and
                  Xiaochen Zuo and
                  Yu Yue and
                  Weinan Dai and
                  Tiantian Fan and
                  Gaohong Liu and
                  Juncai Liu and
                  Lingjun Liu and
                  Xin Liu and
                  Haibin Lin and
                  Zhiqi Lin and
                  Bole Ma and
                  Guangming Sheng and
                  Yuxuan Tong and
                  Chi Zhang and
                  Mofan Zhang and
                  Ru Zhang and
                  Wang Zhang and
                  Hang Zhu and
                  Jinhua Zhu and
                  Jiaze Chen and
                  Jiangjie Chen and
                  Chengyi Wang and
                  Hongli Yu and
                  Yuxuan Song and
                  Xiangpeng Wei and
                  Hao Zhou and
                  Jingjing Liu and
                  Wei{-}Ying Ma and
                  Ya{-}Qin Zhang and
                  Lin Yan and
                  Yonghui Wu and
                  Mingxuan Wang},
  title        = {{DAPO:} An Open-Source {LLM} Reinforcement Learning System at Scale},
  booktitle    = {NeurIPS},
  year         = {2025},

}

@article{DeepSeekMath,
  author       = {Zhihong Shao and
                  Peiyi Wang and
                  Qihao Zhu and
                  Runxin Xu and
                  Junxiao Song and
                  Mingchuan Zhang and
                  Y. K. Li and
                  Y. Wu and
                  Daya Guo},
  title        = {DeepSeekMath: Pushing the Limits of Mathematical Reasoning in Open
                  Language Models},
  journal      = {CoRR},
  volume       = {abs/2402.03300},
  year         = {2024},

}

@inproceedings{yue2025does,
title={Does Reinforcement Learning Really Incentivize Reasoning Capacity in {LLM}s Beyond the Base Model?},
author={Yang Yue and Zhiqi Chen and Rui Lu and Andrew Zhao and Zhaokai Wang and Yang Yue and Shiji Song and Gao Huang},
booktitle={NeurIPS},
year={2025},
}

@inproceedings{IGPO,
title={Information Gain-based Policy Optimization: A Simple and Effective Approach for Multi-Turn Search Agents},
author={Guoqing Wang and Sunhao Dai and Guangze Ye and Zeyu Gan and Wei Yao and Yong Deng and Xiaofeng Wu and Zhenzhe Ying},
booktitle={ICLR},
year={2026},
}

@InProceedings{infopo,
  title = {{I}nfo{PO}: Information-Driven Policy Optimization for User-Centric Agents},
  author = {Kong, Fanqi and Zhang, Jiayi and Deng, Mingyi and Wu, Chenglin and Luo, Yuyu and Liu, Bang},
  booktitle = {Proceedings of the 43rd International Conference on Machine Learning},
  year = {2026},
}

@inproceedings{visual-rtf,
  author       = {Ziyu Liu and
                  Zeyi Sun and
                  Yuhang Zang and
                  Xiaoyi Dong and
                  Yuhang Cao and
                  Haodong Duan and
                  Dahua Lin and
                  Jiaqi Wang},
  title        = {Visual-RFT: Visual Reinforcement Fine-Tuning},
  booktitle    = {ICCV},
  year         = {2025},

}

@article{gpt-4o,
  author       = {OpenAI},
  title        = {GPT-4o System Card},
  journal      = {CoRR},
  volume       = {abs/2410.21276},
  year         = {2024},
}

@inproceedings{Splattalk,
  title={Splattalk: 3d vqa with gaussian splatting},
  author={Thai, Anh and Peng, Songyou and Genova, Kyle and Guibas, Leonidas and Funkhouser, Thomas},
  booktitle={ICCV},
  year={2025}
}

@article{ChatScene,
  title={Chat-scene: Bridging 3d scene and large language models with object identifiers},
  author={Huang, Haifeng and Chen, Yilun and Wang, Zehan and Huang, Rongjie and Xu, Runsen and Wang, Tai and Liu, Luping and Cheng, Xize and Zhao, Yang and Pang, Jiangmiao and others},
  journal={NeurIPS},
  year={2024}
}

@inproceedings{
sqa3d,
title={{SQA}3D: Situated Question Answering in 3D Scenes},
author={Xiaojian Ma and Silong Yong and Zilong Zheng and Qing Li and Yitao Liang and Song-Chun Zhu and Siyuan Huang},
booktitle={ICLR},
year={2023},
}

@InProceedings{Yuan_2025_CVPR,
    author    = {Yuan, Zhihao and Peng, Yibo and Ren, Jinke and Liao, Yinghong and Han, Yatong and Feng, Chun-Mei and Zhao, Hengshuang and Li, Guanbin and Cui, Shuguang and Li, Zhen},
    title     = {Empowering Large Language Models with 3D Situation Awareness},
    booktitle = {CVPR},
    year      = {2025},
}

@article{Qwen2.5-VL,
  title={Qwen2.5-VL Technical Report},
  author={Bai, Shuai and Chen, Keqin and Liu, Xuejing and Wang, Jialin and Ge, Wenbin and Song, Sibo and Dang, Kai and Wang, Peng and Wang, Shijie and Tang, Jun and Zhong, Humen and Zhu, Yuanzhi and Yang, Mingkun and Li, Zhaohai and Wan, Jianqiang and Wang, Pengfei and Ding, Wei and Fu, Zheren and Xu, Yiheng and Ye, Jiabo and Zhang, Xi and Xie, Tianbao and Cheng, Zesen and Zhang, Hang and Yang, Zhibo and Xu, Haiyang and Lin, Junyang},
  journal={arXiv preprint arXiv:2502.13923},
  year={2025}
}

@article{Qwen3-VL,
      title={Qwen3-VL Technical Report}, 
      author={Shuai Bai and Yuxuan Cai and Ruizhe Chen and Keqin Chen and Xionghui Chen and Zesen Cheng and Lianghao Deng and Wei Ding and Chang Gao and Chunjiang Ge and Wenbin Ge and Zhifang Guo and Qidong Huang and Jie Huang and Fei Huang and Binyuan Hui and Shutong Jiang and Zhaohai Li and Mingsheng Li and Mei Li and Kaixin Li and Zicheng Lin and Junyang Lin and Xuejing Liu and Jiawei Liu and Chenglong Liu and Yang Liu and Dayiheng Liu and Shixuan Liu and Dunjie Lu and Ruilin Luo and Chenxu Lv and Rui Men and Lingchen Meng and Xuancheng Ren and Xingzhang Ren and Sibo Song and Yuchong Sun and Jun Tang and Jianhong Tu and Jianqiang Wan and Peng Wang and Pengfei Wang and Qiuyue Wang and Yuxuan Wang and Tianbao Xie and Yiheng Xu and Haiyang Xu and Jin Xu and Zhibo Yang and Mingkun Yang and Jianxin Yang and An Yang and Bowen Yu and Fei Zhang and Hang Zhang and Xi Zhang and Bo Zheng and Humen Zhong and Jingren Zhou and Fan Zhou and Jing Zhou and Yuanzhi Zhu and Ke Zhu},
	  journal={arXiv preprint arXiv:2511.21631},
      year={2025}
}

@article{gemma3,
  author       = {{Gemma Team}},
  title        = {Gemma 3 Technical Report},
  journal      = {CoRR},
  volume       = {abs/2503.19786},
  year         = {2025},

}

@inproceedings{
llava,
title={Visual Instruction Tuning},
author={Haotian Liu and Chunyuan Li and Qingyang Wu and Yong Jae Lee},
booktitle={NeurIPS},
year={2023},
}

@inproceedings{scannet,
  title={Scannet: Richly-annotated 3d reconstructions of indoor scenes},
  author={Dai, Angela and Chang, Angel X and Savva, Manolis and Halber, Maciej and Funkhouser, Thomas and Nie{\ss}ner, Matthias},
  booktitle={CVPR},
  year={2017}
}

@inproceedings{3Rscan,
    title={RIO: 3D Object Instance Re-Localization in Changing Indoor Environments},
  author={Wald, Johanna and Avetisyan, Armen and Navab, Nassir and Tombari, Federico and Nie{\ss}ner, Matthias},
    booktitle={ICCV},
    year={2019}
  }

@inproceedings{arkitscenes,
  author       = {Afshin Dehghan and
                  Gilad Baruch and
                  Zhuoyuan Chen and
                  Yuri Feigin and
                  Peter Fu and
                  Thomas Gebauer and
                  Daniel Kurz and
                  Tal Dimry and
                  Brandon Joffe and
                  Arik Schwartz and
                  Elad Shulman},
  title        = {ARKitScenes: {A} Diverse Real-World Dataset For 3D Indoor Scene Understanding
                  Using Mobile {RGB-D} Data},
  booktitle    = {Proceedings of the Neural Information Processing Systems Track on
                  Datasets and Benchmarks 1, NeurIPS Datasets and Benchmarks 2021, December
                  2021, virtual},
  year         = {2021},

}

\newpage

\appendix

\section{Additional Experimental and Implementation Details}
\label{app:implementation}

\subsection{Experimental Setup}
\label{app:experimental_setup}

\paragraph{Datasets.}  Following prior situated 3D reasoning studies~\cite{msqa,scenecot,Beacon3D}, CARE-3D is evaluated on the widely used MSQA and Beacon3D benchmarks. MSQA is a large scale situated 3D question answering benchmark constructed from ScanNet, 3RScan, and ARKitScenes~\cite{scannet,3Rscan,arkitscenes}. Each question is associated with a 3D scene and a specific observer viewpoint, requiring semantic, geometric, and egocentric reasoning. Following SceneCOT, performance is reported for six question categories: counting, existence, attribute/description, spatial/referential, navigation, and others.

Beacon3D evaluates 3D grounding, question answering, and grounding answer coherence using per case and object level metrics. Supervised fine-tuning is performed using interaction traces derived from MSQA and GQA3D~\cite{scenecot}, whereas reinforcement post-training is conducted only on the MSQA training split. Beacon3D is used exclusively for evaluation.

Following SceneCOT-185K, the supervised data comprise 145.6K MSQA-based situated reasoning traces and 40K GQA3D-based object-centric traces. Reinforcement post-training uses only the MSQA training split, while evaluation follows the official MSQA and Beacon3D evaluation sets.

MSQA contains 10,891 sets comprising 34,357 samples with evidence augmentation. Within each group, samples share the scene, question, and target but differ in observer pose, with pose-consistent answers. Our SFT baseline retains the corresponding pose-augmented training examples.

\paragraph{Baselines.} The evaluation compares \textsc{CARE-3D} with representative multimodal and 3D vision-language methods, including GPT-4o, SplatTalk, LEO, MSR3D, Chat-Scene, and SceneCOT~\cite{gpt-4o,Splattalk,LEO,msqa,ChatScene,scenecot}. These baselines cover closed source multimodal models, general purpose 3D agents, situated reasoning models, and methods based on object centric or 3D Gaussian scene representations. SceneCOT is the closest grounded reasoning baseline and learns stepwise grounding interactions from supervised traces. 

Two controlled training baselines are further included. The \emph{SFT} baseline corresponds to the policy trained solely with supervised grounding traces. The \emph{Answer+Format} baseline applies GRPO with only answer and format rewards. It excludes CARE and GCS. Other component combinations are examined in the ablation study.

\paragraph{Evaluation Metrics.} Following the official evaluation protocols, GPT-Score is used to measure the semantic correctness of generated answers. For MSQA, performance is reported for six reasoning categories: counting, existence, attribute/description, spatial/referential, navigation, and others. The overall score is computed over all evaluation questions. For Beacon3D, \emph{Case} averages the scores of individual question answer pairs, while \emph{Obj.} requires all questions associated with the same target object to be answered correctly. The latter provides a stricter measure of grounding answer coherence. All metrics are reported on a 0-100 scale, and higher values indicate better performance. The LLM judge used as the answer reward during reinforcement post-training is not used for final evaluation.

\paragraph{Implementation.} CARE-3D uses Qwen3-VL-4B-Instruct as the primary vision-language agent, adapted with LoRA of rank 16 and scaling factor 32. The agent and 3D grounding module are first jointly optimized using supervised interaction traces. Reinforcement post-training starts from the resulting SFT checkpoint, with the grounding module and vision-language backbone frozen and only the agent’s LoRA parameters updated. GRPO optimization is conducted on MSQA for 600 steps, using 24 prompts and 192 sampled trajectories per step. Each GRPO group contains six free rollouts and two GCS coverage rollouts. The answer reward is computed using an LLM judge.

\subsection{Training Configurations}

\label{app:training_configurations}

Supervised fine-tuning jointly optimizes the agent and 3D grounding module using supervised demonstrations. GRPO reinforcement post-training starts from the resulting SFT checkpoint, freezes the grounding module and vision-language backbone, and updates only the agent’s LoRA parameters. No separate grounder only refinement is performed. Tables \ref{tab:training_config_sft} and \ref{tab:training_config_grpo} summarize the optimization settings for supervised fine-tuning and reinforcement post-training, respectively.

\begin{table}[!t]
\centering
\footnotesize
\setlength{\tabcolsep}{4pt}
\renewcommand{\arraystretch}{0.95}
\begin{tabular}{@{}p{0.32\columnwidth}p{0.60\columnwidth}@{}}
\toprule
Configuration & Supervised Training \\
\midrule
Base agent & Qwen3-VL-4B-Instruct \\
Trainable params & Agent LoRA, 3D encoder/projector, grounding head \\
Optimizer & AdamW, betas $(0.9, 0.999)$ \\
Learning rate & $2\times10^{-5}$ (agent and grounder, shared group) \\
Weight decay & $0.05$ \\
LR schedule & 400 step linear warm up, cosine decay \\
Duration & 5 epochs \\
Precision & BF16  \\
Hardware & 2$\times$ NVIDIA H100 80GB \\
\bottomrule
\end{tabular}
\caption{Supervised fine-tuning configuration for jointly optimizing the agent adapters and grounding module.}
\label{tab:training_config_sft}
\end{table}

\begin{table}[!t]
\centering
\footnotesize
\setlength{\tabcolsep}{4pt}
\renewcommand{\arraystretch}{0.95}

\begin{tabular}{@{}p{0.32\columnwidth}p{0.60\columnwidth}@{}}
\toprule
Configuration & GRPO Reinforcement Post-Training \\
\midrule
Initialization & SFT checkpoint \\
Trainable params & Agent LoRA only (grounder frozen) \\
Optimizer & AdamW, betas $(0.9, 0.999)$ \\
Learning rate & $1\times10^{-6}$ \\
Weight decay & $0.01$ \\
Duration & 600 GRPO steps  \\
Precision & BF16  \\
Hardware & 2$\times$ NVIDIA H100 80GB \\
\bottomrule
\end{tabular}
\caption{GRPO reinforcement post-training configuration, with the grounding module frozen and only the agent LoRA parameters updated.}
\label{tab:training_config_grpo}
\end{table}

\paragraph{LoRA configuration.}
Supervised fine-tuning and reinforcement post-training use the same LoRA architecture, with rank $r=16$, scaling parameter $\alpha=32$, and an effective scaling ratio of $\alpha/r=2$. LoRA is applied to the
\texttt{q\_proj}, \texttt{k\_proj}, \texttt{v\_proj},
\texttt{o\_proj}, \texttt{gate\_proj}, \texttt{up\_proj}, and
\texttt{down\_proj} modules. During Supervised fine-tuning, the agent LoRA parameters and all trainable components of the grounding module share a single AdamW parameter group and therefore use the same learning rate of $2\times10^{-5}$.

\paragraph{Supervised fine-tuning batch construction.}
Supervised fine-tuning uses four training samples per GPU, two GPUs, and five gradient accumulation steps, resulting in an effective global batch size of $4\times2\times5=40$ samples per optimizer update. The maximum input context length is 1,024 tokens, and the maximum supervised output length is 768 tokens.

\paragraph{GRPO batch construction.}
Batch parameters in GRPO operate at different granularities: the training batch size counts \emph{prompts}, whereas actor updates operate on \emph{trajectories}. Each GRPO step processes 24 prompts, and each prompt yields a group of $G=8$ trajectories
(six free on-policy rollouts and two GCS coverage rollouts):
\[
\scalebox{0.9}{$\displaystyle
\begin{aligned}
B_{\mathrm{rollout}} &= 24\ \text{prompts}\times 8 = 192\ \text{trajectories},\\
B_{\mathrm{mini}}    &= 12\ \text{prompt groups}\times 8 = 96\ \text{trajectories},\\
B_{\mathrm{micro}}   &= 4\ \text{traj.\ per GPU}\times 2\ \text{GPUs} = 8\ \text{trajectories},\\
N_{\mathrm{acc}}     &= B_{\mathrm{mini}}/B_{\mathrm{micro}} = 96/8 = 12.
\end{aligned}
$}
\]
Each GRPO step thus contains $192/96=2$ actor mini batches; with one PPO epoch, 600 GRPO steps correspond to approximately 1{,}200 actor optimizer updates, and the complete run samples $600\times 24\times 8 = 115{,}200$ training trajectories.

\paragraph{GRPO hyperparameters.}
The GRPO clipping parameter is set to $\varepsilon=0.2$ such that the policy ratio is clipped to $[1-\varepsilon,1+\varepsilon]
=[0.8,1.2]$, giving symmetric lower and upper clipping ranges of $0.2$. The coefficient of the KL regularization term in Equation~\ref{eq:grpo_objective} is $\beta=0.02$. KL divergence is estimated using a \texttt{low variance KL estimator} and is added directly to the actor loss; it is not incorporated into the scalar rollout reward. CARE-3D uses one PPO epoch per GRPO step, an entropy coefficient of zero, and a maximum gradient norm of $1.0$.

\paragraph{Rollout generation.}
Training rollouts are sampled with temperature $0.7$, top-$p=0.9$, and no top-$k$ truncation. The maximum prompt length is 2,048 tokens, and the complete single grounding interaction has a total response budget of 1,700 tokens. Validation uses greedy decoding with temperature zero, top-$p=1.0$, and one generated response per question.

\subsection{Reward Configuration}
\label{app:reward_config}
For executable rollouts, the scalar reward follows Equation~\ref{eq:overall_reward} and combines format validity, answer correctness, and CARE-based evidence utility. Catastrophically invalid rollouts, whose grounding requests cannot be parsed or executed, bypass this aggregation and receive a fixed reward of $R_i=-1.0$. Table~\ref{tab:reward_config} summarizes the reward weights and invalid rollout handling.
\begin{table}[t]
\centering
\small

\begin{tabular}{lcc}
\toprule
Component & Steps 1-200 & Steps 201-600 \\
\midrule
Format weight $w_{\mathrm{fmt}}$ & 1.2 & 0.8 \\
Answer weight $w_{\mathrm{ans}}$ & 0.8 & 1.2 \\
CARE weight $w_{\mathrm{CARE}}$ & 0.2 & 0.2 \\
Invalid rollout reward & \multicolumn{2}{c}{$R_i=-1.0$} \\
\bottomrule
\end{tabular}
\caption{Reward weights and invalid rollout handling during reinforcement post-training.}
\label{tab:reward_config}
\end{table}

\paragraph{Base reward.}
For a non-catastrophic rollout, the reward before lower bound clipping is
\begin{equation}
\scalebox{0.9}{$
\widetilde{R}_i = w_{\mathrm{fmt}}r_i^{\mathrm{fmt}} + w_{\mathrm{ans}}r_i^{\mathrm{ans}}
+ w_{\mathrm{CARE}}r_i^{\mathrm{CARE}} - p_i
$},
\label{eq:reward_before_clipping}
\end{equation}
where $r_i^{\mathrm{fmt}}$ evaluates interaction format validity, $r_i^{\mathrm{ans}}$ evaluates answer correctness, and $r_i^{\mathrm{CARE}}$ measures the utility of returned grounded evidence. The final reward is
\begin{equation}
\scalebox{0.9}{$
R_i = \begin{cases} -1.0, & \text{if $\tau_i$ is invalid},\\ \max\!\left(\widetilde{R}_i,\,-1.0\right),
& \text{otherwise}.
\end{cases}
$}
\label{eq:final_reward}
\end{equation}
Catastrophically invalid responses include outputs whose grounding calls cannot be parsed or executed. Assigning a fixed reward of $-1.0$ prevents such trajectories from receiving positive credit from other reward components.

\paragraph{Format and answer rewards.} The format reward checks whether the response follows the required grounding interaction protocol. Unless otherwise specified, the answer reward is computed by Gemini-2.5-Flash-Lite, which evaluates the semantic consistency between the generated and ground truth answers. Let $s_i^{\mathrm{Gemini}} \in \{1,2,3,4,5\}$ denote the integer score returned by the judge. This score is linearly normalized to obtain the answer reward:
\begin{equation}
r_i^{\mathrm{ans}}
=
\frac{s_i^{\mathrm{Gemini}}-1}{4}.
\end{equation}
Therefore, $r_i^{\mathrm{ans}} \in
\{0, 0.25, 0.5, 0.75, 1\}$, where a clearly incorrect answer
receives zero answer reward and a fully correct answer receives
the maximum answer reward. Additional thresholding or binarization is not applied. The complete judge prompt is provided in Appendix~\ref{app:prompts}. This training time judge is separate from the GPT-Score evaluator used for final benchmark evaluation.

\paragraph{Sensitivity to the answer reward judge.} We examine the sensitivity of CARE-3D to the language model
used to compute the answer reward. In addition to the default Gemini-2.5-Flash-Lite judge, we use Qwen3-4B and GPT-5.6 Luna, while keeping all other training settings and the final benchmark evaluator fixed. Table~\ref{tab:reward_judge} shows score differences of at most 0.7, 0.6, and 0.4 points on MSQA Overall, Beacon3D Case, and Beacon3D Obj., respectively. These results suggest limited sensitivity to the answer reward judge among the three tested models, including the open-source Qwen3-4B.

\begin{table}[h]
    \centering
    \footnotesize
    \setlength{\tabcolsep}{3pt}
    \begin{tabular}{lccc}
        \toprule
        \multirow{2}{*}{Answer reward judge}
        & MSQA & \multicolumn{2}{c}{Beacon3D} \\
        \cmidrule(lr){2-2}
        \cmidrule(lr){3-4}
        & Overall$\uparrow$ & Case$\uparrow$ & Obj.$\uparrow$ \\
        \midrule
        Qwen3-4B             & 65.1 & 63.9 & 25.3 \\
        GPT-5.6 Luna         & 65.8 & 64.3 & 25.6 \\
        Gemini-2.5-Flash-Lite & 65.7 & 64.5 & 25.7 \\
        \bottomrule
    \end{tabular}
    \caption{Sensitivity of CARE-3D to the answer reward
    judge used during reinforcement post-training.
    All variants use the same final benchmark evaluator.}
    \label{tab:reward_judge}
\end{table}

\paragraph{CARE reward.} The CARE reward is computed according to Equation~\ref{eq:care_reward}. The factual and evidence masked likelihoods are evaluated using the fixed rollout policy $\pi_{\theta}$. The resulting reward lies in $(0,1)$ and is assigned a constant weight of $w_{\mathrm{CARE}}=0.2$ throughout reinforcement post-training.

\paragraph{Reward schedule and controlled variants.} For the first 200 optimization steps, the format and answer weights are set to $1.2$ and $0.8$, respectively. They are switched to $0.8$ and $1.2$ for the remaining 400 steps. The reward schedule and answer judge are shared across all reinforcement post-training variants. Variants without CARE set $w_{\mathrm{CARE}}=0$, whereas GCS changes only rollout group construction and introduces no additional reward.

\subsection{CARE Implementation}
\label{app:care_implementation}

CARE follows Equations~\ref{eq:care_delta}
and~\ref{eq:care_reward}. It is computed after rollout generation and before group relative advantage estimation. Only executable rollouts are scored.

\paragraph{Likelihood computation.}
The policy used to generate the rollouts also serves as the CARE scoring policy. It is placed in evaluation mode and remains fixed during reward computation. For each factual or evidence masked context, CARE computes the mean log likelihood over the non padding ground truth answer tokens. Prompt tokens, returned evidence, and padding tokens are excluded from this average. The two contexts score exactly the same ground truth answer tokens, and no gradient is propagated through either likelihood computation.

\paragraph{Normalization and GCS integration.}
Evidence Gains are normalized across the executable rollouts associated with the same question. The normalization constant is set to $\epsilon=10^{-6}$ for numerical stability. The same scoring procedure is applied to free and GCS coverage rollouts. For a coverage rollout, the tool identifier inserted by GCS remains in both scoring contexts but is excluded from the policy loss. CARE therefore evaluates the returned evidence without treating the inserted tool identifier as a supervised target.

\subsection{GCS Implementation}
\label{app:gcs_implementation}
Group Coverage Sampling (GCS) is applied independently to each training prompt during reinforcement post-training. The policy first samples six free on-policy rollouts using the rollout generation settings reported in Appendix~\ref{app:training_configurations}. The grounding tool identifier in each valid rollout is parsed, and the occurrence count of each tool in $\{\texttt{Prob},\texttt{Cartesian},\texttt{Polar},\texttt{Crop}\}$ is recorded. Malformed calls and unsupported tool identifiers do not increase the count of any valid tool.

The two tools with the lowest occurrence counts are selected for coverage sampling. One intervened rollout is generated for each selected tool, producing a group of eight rollouts: six free rollouts and two intervened coverage rollouts. In a coverage rollout, GCS supplies only the tool identifier. The grounding query, reasoning trace, and final answer remain generated by the current policy, and the grounding module returns evidence normally using the supplied tool.

A token level loss mask excludes each GCS supplied tool identifier from the policy loss. Returned tool evidence is also excluded because it is generated by the grounding module rather than the policy. All remaining policy generated tokens participate normally in GRPO optimization. Free and coverage rollouts use the same reward function, CARE computation, and group relative advantage estimation. GCS therefore changes the composition of the rollout group without introducing an additional reward or training the policy to imitate externally selected tools.

GCS is enabled only during reinforcement post-training and is disabled during evaluation, where the policy selects grounding tools without GCS coverage sampling. The default number of coverage rollouts is set to two. The effect of varying the number of coverage is reported in Table~\ref{tab:gcs_sensitivity}.

\subsection{Error Analysis and Evidence Interventions}
\label{app:evidence_analysis}

A correct answer alone does not establish appropriate evidence use. We therefore conduct error-type and evidence-intervention analyses on 1,369 validation samples to compare the SFT and CARE-3D policies.

\paragraph{Error-type analysis.}

We examine two diagnostic error types: incorrect grounding and misinterpretation of returned evidence. As shown in Table~\ref{tab:evidence_errors}, CARE-3D improves answer accuracy from 49.7\% to 56.8\%, while reducing incorrect grounding from 37.3\% to 29.0\% and misinterpreted evidence from 9.8\% to 4.5\%. These results suggest improvements in both evidence acquisition and interpretation.

\begin{table}[h]
\centering
\small
\setlength{\tabcolsep}{4pt}
\begin{tabular}{lccc}
\hline
Policy &
Acc.$\uparrow$ &
\shortstack{Incorrect\\Grounding$\downarrow$} &
\shortstack{Misinterpreted\\Evidence$\downarrow$} \\
\hline
SFT  & 49.7\% & 37.3\% & 9.8\% \\
CARE-3D & 56.8\% & 29.0\% & 4.5\% \\
\hline
\end{tabular}
\caption{Answer accuracy and diagnostic error rates on 1,369 validation samples.}
\label{tab:evidence_errors}
\end{table}

\subsection{Software and licenses.}
Our implementation builds on the Apache-2.0-licensed SceneCOT and
\texttt{verl} codebases. We use \texttt{verl} at commit
\texttt{93c3dd1daba7a836b3fae58cd9d17aea4f55ae9f}. Final evaluation follows the official MSQA and Beacon3D GPT-Score protocols using \texttt{gpt-4o-2024-11-20}, with temperature~0 and one evaluation call per predicted answer. The released code retains the applicable license and attribution notices and does not redistribute third-party datasets or pretrained model weights.

\section{Prompts}
\label{app:prompts}

This section presents the prompts used for LLM based answer evaluation. In the implementation prompts, \emph{observation} refers to the grounded evidence returned by the selected tool.

\begin{promptbox}{Box 2: Training Time Answer Judge Prompt}
Score open-ended answers from 1 to 5 based on semantic accuracy, completeness, and relevance to the ground truth.

Please follow the following rules:
- Judge semantic correctness, not exact wording.
- Accept synonyms, paraphrases, singular/plural changes, harmless extra details, and reasonable specificity differences.
- If the prediction is more general than the ground truth but captures the main answer, give partial credit.
- For object lists, do not require every item unless the question asks for all objects.
- For spatial or navigation answers, minor direction or distance errors should receive partial credit.
- Only give Score 1 when the answer is clearly wrong, contradicts the ground truth, is irrelevant, or refuses to answer when the ground truth provides an answer.
\end{promptbox}

\section{Qualitative Examples}
\label{app:qualitative}

This section presents qualitative comparisons between the SFT policy and the CARE-3D policy across three representative reasoning categories: counting, existence, and navigation. Each example compares the grounding request, selected tool, returned scene evidence, and final prediction produced by the two policies. These cases illustrate how reinforcement post-training enables CARE-3D to select more task appropriate grounding tools and make better use of the returned evidence. In particular, CARE-3D exhibits stronger grounding answer coherence, producing answers that are more consistently supported by the acquired visual and spatial information.
\begin{figure}[H]
    \centering
    \includegraphics[width=\columnwidth]
    {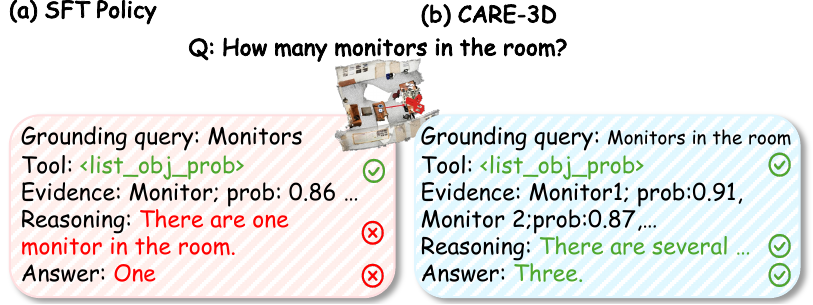}
    \caption{Qualitative comparison on a counting question. CARE-3D uses a more specific grounding query to retrieve sufficient object evidence and produce the correct count.}
    \label{fig:case_counting}
\end{figure}

\begin{figure}[H]
    \centering
    \includegraphics[width=\columnwidth]
    {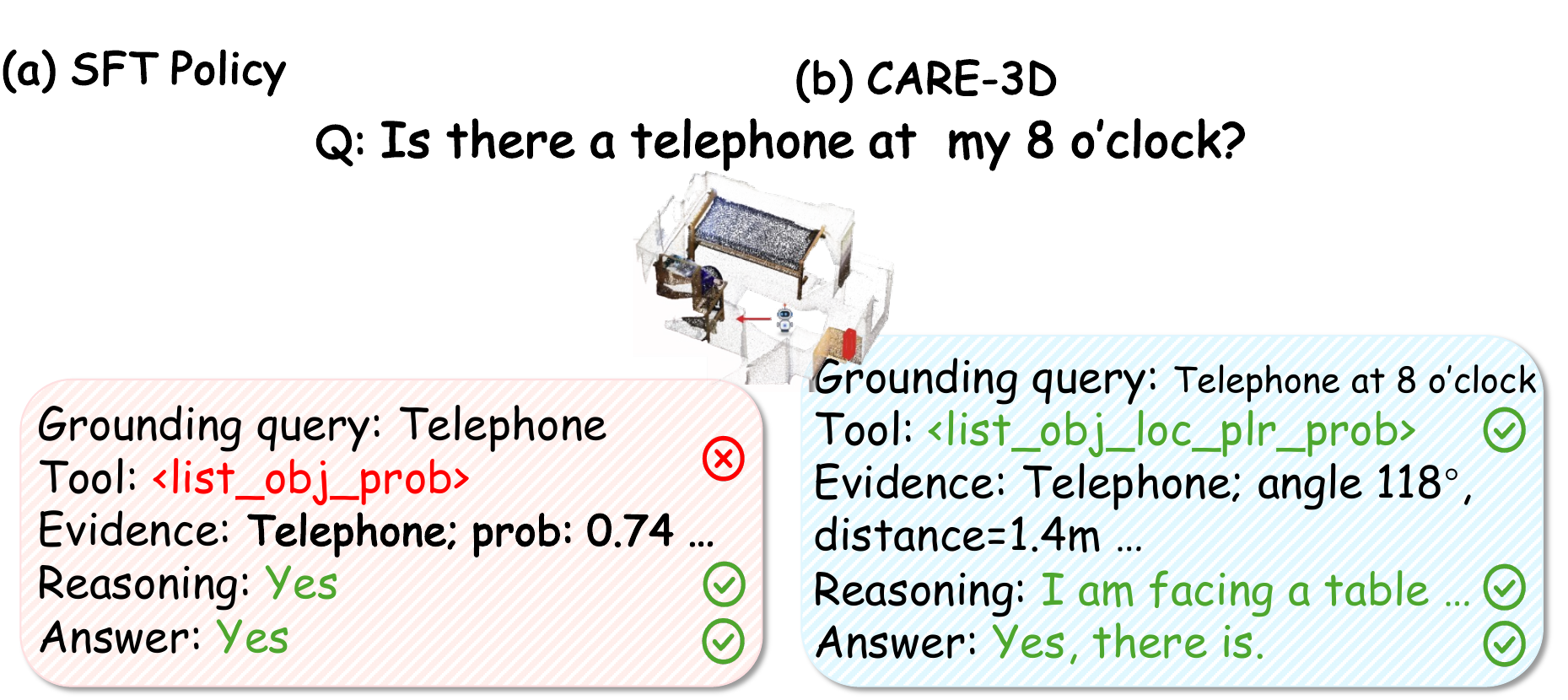}
    \caption{Qualitative comparison on an existence question. CARE-3D selects viewer relative polar evidence that directly supports the specified clock direction.}
    \label{fig:case_existence}
\end{figure}

\begin{figure}[H]
    \centering
    \includegraphics[width=\columnwidth]
    {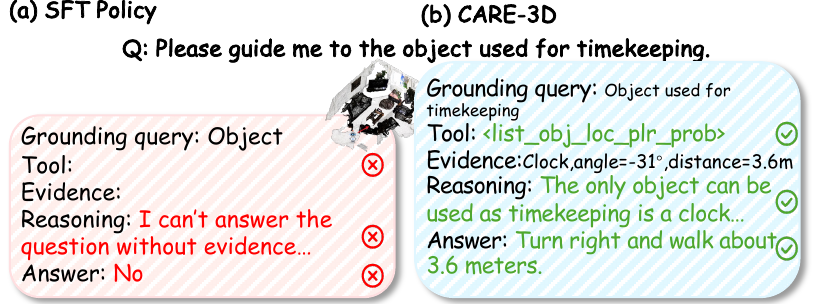}
    \caption{Qualitative comparison on a navigation question. CARE-3D retrieves viewer relative direction and distance evidence to generate a grounded navigation instruction.}
    \label{fig:case_navigation}
\end{figure}

\end{document}